\documentclass[runningheads]{llncs}

\usepackage{eccv}

\usepackage{eccvabbrv}
\usepackage{float}

\usepackage{wrapfig}

\usepackage{graphicx}
\usepackage{booktabs}
\usepackage{gensymb}
\usepackage{booktabs}
\usepackage{multirow}

\usepackage[accsupp]{axessibility}  % Improves PDF readability for those with disabilities.

\newcommand{\methodname}{3D FaceShell\xspace}
\usepackage[table]{xcolor}
\definecolor{low}{RGB}{255,200,200}
\definecolor{mid}{RGB}{255,255,200}
\definecolor{high}{RGB}{220,240,220}

\usepackage{hyperref}

\usepackage{orcidlink}

\begin{document}

% \title{3D Face Attribute Transfer for Vision–Language Model Defense } 
\title{\methodname: Attribute Transfer in 3D Face Avatars as a VLM Defense Mechanism}

\titlerunning{\methodname}

\author{Weston Bondurant \and Srijan Das\setcounter{footnote}{3}\thanks{Equal advising.} \and Hieu Le\protect\footnotemark[4] \and Stephanie Schuckers\protect\footnotemark[4]}

\authorrunning{W.~Bondurant et al.}

\institute{University of North Carolina at Charlotte \\
\email{wbondura@charlotte.edu} \\
\url{https://westonbond.github.io/3DFS/}}

\maketitle
% \vspace{-0.4pt}
\begin{figure}[h]
    \centering
    \scalebox{0.8}{
    \includegraphics[width=1\linewidth]{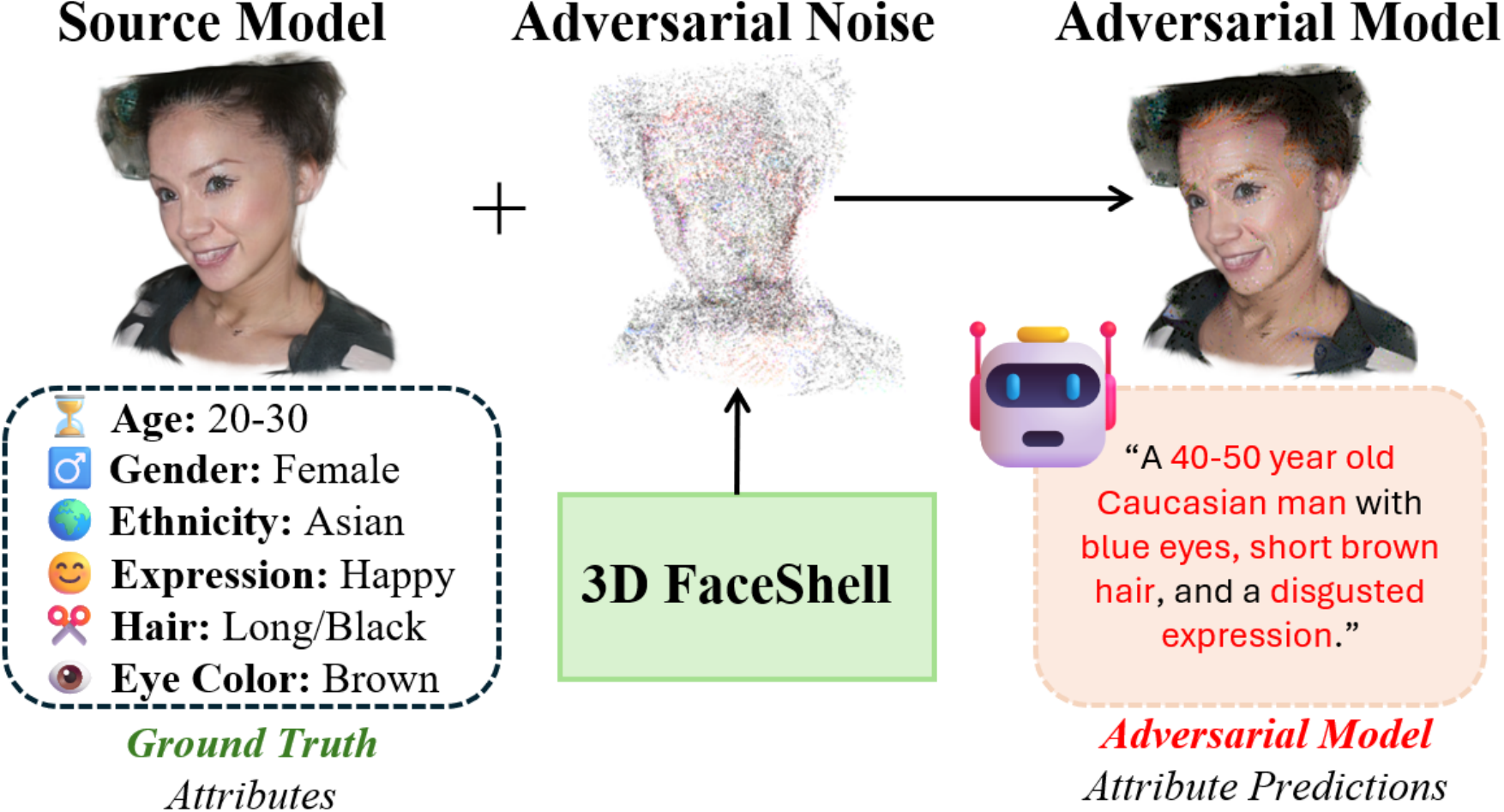}}
    \caption{Illustration of our 3D FaceShell method. By optimizing a minimally perceptible 3D Gaussian shell, our method successfully subverts VLM attribute predictions across rendered views while preserving the original facial identity and photorealistic appearance.} \vspace{-0.5in}
    \label{Teaser}
\end{figure}

\begin{abstract}
% Privacy concerns are increasing as high-fidelity 3D face representations become easier to capture and share. While prior work has largely focused on protecting rendered images, comparatively little attention has been given to safeguarding the 3D representations themselves—an emerging but increasingly important form of visual data. We study privacy protection directly at the representation level, treating the 3D asset as the unit of defense so that rendered views inherit protective properties by construction.

% We instantiate this idea by augmenting a 3D face representation (e.g., a Gaussian-splat model) with learnable auxiliary splats embedded in the same space. These splats are placed near the facial surface and optimized to remain visually inconspicuous while influencing signals used by downstream models. Because they are additive, the original geometry and identity-defining structure remain unchanged, preserving both identity and visual quality in rendered views.

% Experiments with vision–language models show that this representation-level augmentation reduces their ability to infer facial attributes while preserving visual fidelity. These results suggest that as 3D representations become more widely adopted, designing protection mechanisms at the same level offers a practical alternative to image-space defenses.

Photorealistic 3D face avatars are increasingly deployed as reusable digital assets across applications such as telepresence, animation, and personalized media. At the same time, vision–language models (VLMs) can infer sensitive attributes from rendered images with open-ended semantic reasoning without any fine-tuning. This creates a new privacy challenge: once a 3D face avatar is shared, any of its renderings can be analyzed to extract high-level facial attributes. Existing defenses largely operate in 2D image space and do not address identity-preserving semantic manipulation of 3D facial representations.

We propose \textbf{\methodname}, a framework for steering VLM interpretations of faces rendered from 3D models while preserving geometric fidelity and facial identity. \methodname~augments the original 3D representation with a learnable Gaussian shell that produces subtle, spatially distributed perturbations optimized through multi-view embedding alignment. The perturbations are designed to be visually inconspicuous yet sufficient to redirect VLM-based attribute inference in a view-consistent manner.

Extensive experiments on reconstructed celebrity face avatars and multiple black-box VLMs demonstrate that \methodname~significantly increases attribute injection and mismatch rates while maintaining high perceptual similarity and identity consistency. Our results show that it is possible to manipulate VLM-level semantic interpretation of 3D faces without compromising their human-recognizable appearance.
\keywords{3D \and Gaussian Splatting \and Vision–Language Models \and Face Attribute Inference}
\end{abstract}
\vspace{-0.2in}
\vspace{-0.2in}
\section{Introduction}
\label{sec:intro}
Photorealistic 3D face avatars are rapidly becoming reusable digital assets. Modern pipelines reconstruct detailed 3D heads from short videos or even single images, enabling talking-head synthesis, virtual telepresence, personalized media, and immersive applications\cite{facelift, gaussianavatars, hravatar}. Unlike a static photograph, a 3D face model is persistent: it can be rendered under arbitrary viewpoints and lighting conditions long after its creation. At the same time, Vision–Language Models (VLMs), including both commercial foundation models such as GPT-4 or Gemini \cite{gpt, gemini} and open source models like VideoLLaMA3 or LLaVa-NeXT \cite{VL3, llavanext}, have evolved into open-ended semantic reasoners capable of inferring attributes such as \textit{age}, \textit{ethnicity}, \textit{gender}, and \textit{expression} from any rendered image without task-specific training\cite{liu2023_llava, meta2024llama3herdmodels}. Together, these developments create a structural privacy risk: once a 3D face avatar is released, every possible rendering becomes analyzable by powerful semantic models, exposing sensitive attributes beyond simple identity recognition~\cite{facellava}.

Existing defenses are fundamentally mismatched to this setting. Most face anonymization and adversarial cloaking methods operate in 2D image space, protecting only a specific rendering rather than the underlying 3D asset. A new viewpoint instantly bypasses such defenses. While recent adversarial work has explored perturbations in 3D representations, these approaches primarily target identity detectors and do not address identity-preserving semantic manipulation in VLM embedding space \cite{physicalsurvey, effectiveadversarialtexture3d, facerecfacespoof, physicalworld}.

To this end, we propose \textbf{\methodname}, a framework that steers VLM interpretations of 3D face avatars while preserving the underlying facial identity and appearance as highlighted in Fig. \ref{Teaser}. Our objective is to \textit{alter specific attributes} inferred by VLMs from a 3D face avatar while preserving the underlying facial identity and geometry. Unlike generic 3D objects, facial avatars are designed to maintain a stable and recognizable digital likeness for legitimate uses such as telepresence, animation, and personalized media. Any intervention must therefore balance two competing requirements: it must be strong enough to redirect high-level semantic interpretation in the VLM embedding space, yet subtle enough to preserve geometric structure and identity-defining appearance across viewpoints. Naive perturbations that visibly distort shape or degrade texture undermine the asset’s utility. The challenge is thus not merely to attack a model, but to manipulate semantic inference in a way that remains perceptually faithful and multi-view consistent.

To realize this objective, \methodname~operates directly in the parameter space of 3D Gaussian Splatting (3DGS) face avatars. Starting from a pre-optimized facial representation, we introduce a set of auxiliary Gaussians positioned near the facial surface, forming a learnable Gaussian shell. The baseline Gaussians remain fixed, separating adversarial perturbations from underlying geometry. This approach enables semantic manipulation while, by design, never deforming facial structure. The auxiliary shell is optimized to produce subtle, spatially distributed perturbations that are minimally perceptible in rendered views yet sufficient to shift VLM embeddings toward a targeted semantic interpretation. During training, multiple rendered views are passed through a pretrained VLM, and the shell parameters are updated via multi-view embedding alignment against a pose-consistent, attribute-altered target. This design ensures that the semantic manipulation is both view-consistent and visually inconspicuous.

Through extensive evaluation on reconstructed celebrity face avatars and multiple opaque-box VLMs, we show that \methodname~consistently increases attribute injection and mismatch rates while maintaining high perceptual similarity and identity consistency. Compared to state-of-the-art 2D semantic attack baselines, \methodname~achieves stronger cross-view robustness and substantially better preservation of facial structure, demonstrating that semantic steering and visual fidelity need not be mutually exclusive.

We summarize our contributions as follows:
\begin{itemize}
    \item We introduce and formalize \textbf{the problem of manipulating VLM perception of semantic face attributes for 3D face avatars} under strict identity and geometric preservation constraints. Defending against VLM inference for 3D avatars is entirely new, differing fundamentally from prior 3D adversarial attacks that target face detectors and more general adversarial attacks on VLMs that do not preserve face quality or identity.
    \item We propose \textbf{\methodname, a novel additive Gaussian shell framework} that injects spatially distributed, minimally perceptible perturbations into 3DGS face representations and optimizes them through multi-view embedding alignment in the vision encoder space to manipulate VLM perception. This design enables view-consistent semantic steering while leaving the baseline geometry untouched.
    \item We demonstrate \textbf{strong cross-view robustness and improved identity preservation} compared to state-of-the-art 2D semantic attack baselines across multiple opaque-box VLMs, showing that semantic manipulation and perceptual fidelity can be achieved simultaneously in 3D.
\end{itemize}

\vspace{-0.2in}
\section{Related Work}
\vspace{-0.1in}
\subsection{3D Gaussian Splatting}
The introduction of 3DGS has established a highly efficient paradigm for real-time radiance field rendering\cite{3dgs}. By utilizing millions of 3D Gaussians characterized by learnable opacity, scale, rotation, and spherical harmonics, 3DGS achieves state-of-the-art view synthesis without the massive computational bottlenecks of earlier methods \cite{3dgs_advances, mip, lightgaussian, 4d}. Given this explicit editability and rendering speed, 3DGS has been widely adopted for modeling dynamic and highly detailed human face and body avatars. Recent frameworks \cite{facelift, gaussianavatars, hravatar} enable the generation of full body or face avatars from 2D video or images. However, as these 3D head avatars become increasingly common and easily shareable digital assets, the underlying geometry and identity textures are left inherently exposed to malicious downstream processing and automated semantic parsing by VLMs\cite{privacyavatar,privateattribute,facexbench,exploringvlms}. This danger necessitates privacy defenses for 3D face avatars. 
\vspace{-0.2in}
\subsection{Adversarial Attacks on 3D Models}

As the use of 3D representations has become more common, the vulnerabilities of 3D representations to adversarial manipulation have become a critical security domain \cite{mvcamo, advsurvey}. Discrete 3D structures, such as point clouds, are highly susceptible to targeted perturbations either through point shifting or malicious point generation \cite{advpointcloud}. However, adversarial attacks in the 3D domain are not limited to explicit point sets. Implicit representations like Neural Radiance Fields (NeRFs) are vulnerable to targeted perturbations that hallucinate or obscure objects during view synthesis\cite{targetednerf}. Recently, this threat has been explicitly extended to 3DGS. Recent works inject adversarial noise in the 3DGS space to fool object detectors or vision encoders \cite{3dgsunderattack, mvcamo}. Unlike these methods which shift overall semantic representations in object detectors or vision encoders for objects, we aim instead to shift VLM understanding of a more complex domain, \textit{faces}, targeting specific attributes for transfer rather than broader semantics of the 3D representation.

\textbf{Attacks on 3D Face Avatars.} Biometric authentication systems are very frequently targeted by adversarial attacks leading to the development of specialized physical and 3D face attacks that bypass conventional defenses \cite{physicalsurvey, effectiveadversarialtexture3d, facerecfacespoof, physicalworld}. These methods, however, focus on attacks utilizing physical 3D masks or traditional methods of 3D representation like point clouds. In contrast, our \methodname~proposes an attack on 3DGS face avatars, utilizing exclusively additive noise which leaves the underlying face representation undisturbed. 
\vspace{-0.2in}
\subsection{Adversarial Attacks on VLMs}

Due to their extensive cross-modal training, foundation VLMs possess powerful reasoning capabilities but remain highly susceptible to visual adversarial attacks\cite{visualadversarial}. Images with adversarial noise applied to them can bypass safety guardrails and execute multimodal jailbreaks, coercing models into generating harmful or unintended text \cite{jailbreakingmllm, robustmultimodal}. The threat landscape includes both targeted attacks on specific commercial systems \cite{ssacwa} along with targeted and untargeted transferable frameworks designed to disrupt generalized vision-language alignment across a variety of VLMs \cite{vattack, anyattack, attackvlm, advdiff, mattack}. \methodname~differs from these by moving the problem of attacking VLMs with adversarial noise into the 3D space and focusing on shifting the attributes of a face representation while still maintaining a high level of detail and realism to retain the perceptual and identity similarity relative to the source face.
\vspace{-0.1in}
\section{Problem Definition}
\vspace{-0.1in}
\label{problem_definition}

We represent a 3D face using 3DGS as a set of $N$ anisotropic Gaussians 
$\mathcal{G} = \{g_i\}_{i=1}^{N}$, where each Gaussian $g_i$ is parameterized by its spatial mean 
$\mathbf{x}_i \in \mathbb{R}^3$, covariance $\Sigma_i \in \mathbb{R}^{3 \times 3}$, and opacity 
$\alpha_i \in \mathbb{R}$. Let $R(\mathcal{G}; \pi)$ denote the differentiable renderer that produces a 2D image from $\mathcal{G}$ under camera parameters $\pi$. 
Simultaneously, we consider a pretrained VLM $f_{\theta}(\cdot)$, composed of a visual encoder $\mathrm{VE}(\cdot)$ followed by a large language model $\mathrm{LLM}(\cdot)$:
$
f_{\theta}(x) = \mathrm{LLM}(\mathrm{VE}(x)),
$
where $\theta$ denotes fixed model parameters.

Then, given a source 3D face $\mathcal{G}_s$ with ground-truth semantic description $c_{gt}$, our objective is to learn an additive perturbation $\delta$ in the Gaussian parameter space, yielding a modified 3D representation $\mathcal{G}_t = \mathcal{G}_s + \delta$, such that rendered views $R(\mathcal{G}_t; \pi)$ are semantically aligned with a predefined target description $\mathcal{T}_{tar}$ (targeted attack), while preserving the original facial identity and geometric structure.
Under a white-box setting with access to $f_\theta$, the optimization problem is formulated as:
\begin{equation}
\min_{\delta} \;
\mathcal{L}_{adv}\big(f_\theta(R(\mathcal{G}_s + \delta; \pi)), \mathcal{T}_{tar}\big)
+ \lambda \, \mathcal{L}_{reg}(\mathcal{G}_s, \mathcal{G}_s + \delta),
\label{objective}
\end{equation}
where $\mathcal{L}_{adv}$ drives semantic alignment with the target description in the VLM output space, and $\mathcal{L}_{reg}$ constrains the perturbation to preserve geometric fidelity, identity consistency, and photorealistic rendering quality.
 %The key challenge lies in introducing imperceptible yet semantically effective perturbations in the entangled 3DGS parameter space, such that adversarial effects transfer consistently across rendered viewpoints without degrading structural fidelity.

\vspace{-0.2in}

\section{Method: \methodname}
\vspace{-0.1in}
We propose \textbf{\methodname}, an identity-preserving framework for learning additive adversarial perturbations directly in the parameter space of 3DGS facial representations to induce targeted semantic misalignment in pretrained VLMs.
\begin{wrapfigure}{r}{0.48\textwidth}
\vspace{-0.2in}
\centering
\includegraphics[width=\linewidth]{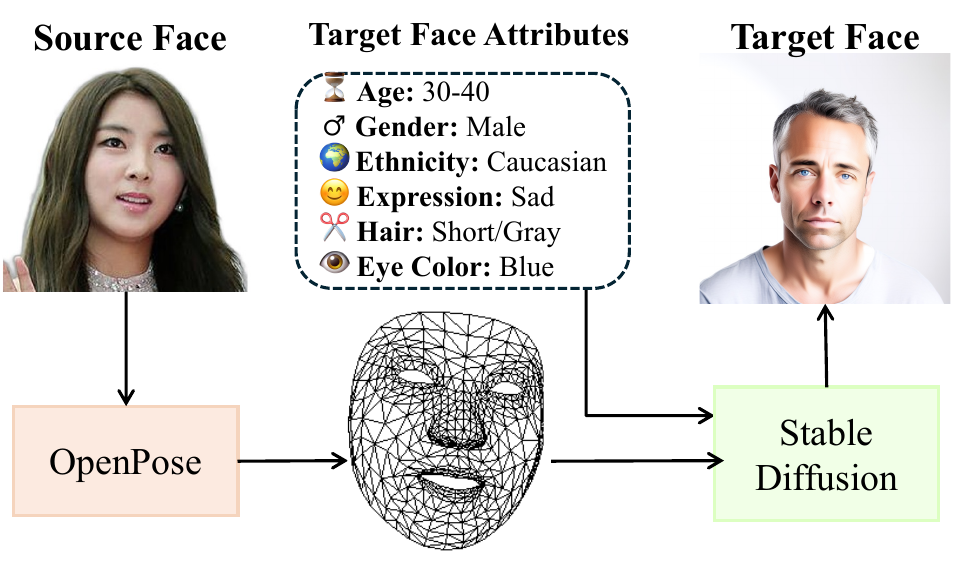}
\caption{Multi-Attributed Target Image Generation. (1) OpenPose extracts the structural pose from a source face. (2) A Stable Diffusion model, conditioned on this pose and a target attribute text prompt, synthesizes a pose-aligned, semantically different target image.}
\vspace{-0.15in}
\label{Target}
\end{wrapfigure}

%Our goal is to induce targeted semantic shifts in pretrained VLMs by optimizing perturbations directly in the 3D Gaussian parameter space, while maintaining the geometric integrity and identity of the underlying face. Unlike 2D adversarial approaches, \methodname\ enforces multi-view consistency by operating on the full 3D representation and propagating gradients through a differentiable renderer.
The proposed framework consists of four key components: 
(a) \emph{Multi-attributed target image generation}, which synthesizes pose-aligned target faces with semantically opposing attributes to guide embedding manipulation; 
(b) \emph{Gaussian shell initialization}, where a secondary set of auxiliary Gaussians is introduced to parameterize controlled additive perturbations without modifying the baseline geometry; 
(c) \emph{Multi-view adversarial training}, which aligns rendered views of the perturbed 3D model with target embeddings under stochastic augmentations for robustness; 
and 
(d) \emph{Face-aware regularization mechanisms}, which constrain perturbations in identity-sensitive regions and suppress unnatural luminance artifacts. 
Together, these components enable semantically effective yet structurally consistent adversarial manipulation of 3D faces in VLM embedding space. 
\vspace{-1.25em}
\subsection{Multi-Attributed Target Image Generation}

To construct supervision for the adversarial objective defined in Sec.~\ref{problem_definition}, we synthesize target images that reflect the desired semantic manipulation $c_{tar}$. Given an input rendering $R(\mathcal{G}_s, \pi)$, we first extract its pose representation using OpenPose\cite{openpose}, producing a structural pose map $\mathbf{P}$ that preserves geometric configuration while disentangling appearance. We then condition a pretrained text-to-image diffusion model on $\mathbf{P}$ together with a text prompt $\mathcal{T}'_{tar}$ encoding a set of desired facial attributes. This process is illustrated in Fig. \ref{Target}. The target prompt is constructed using a multi-attribute composition strategy, where selected attributes are deliberately chosen to be semantically different to those of the original face. While the user may explicitly specify a subset of attributes to modify, the remaining facial attributes are automatically initialized with values opposite to the source description. This multi-attribute formulation encourages a stronger shift in the VLM embedding space, as aligning multiple semantic factors simultaneously produces a more pronounced deviation toward $\mathcal{T}_{tar}$ compared to single-attribute manipulation, analogous to multi-task supervision.
Finally, the synthesized image $I_{tar}$ serves as a semantic anchor for guiding the optimization of the Gaussian shell parameters.

Thus, we reformulate the VLM text-injection objective into a target image injection objective, replacing purely text-driven alignment with embedding-space supervision via a synthesized target image. If $x = R(\mathcal{G}_s; \pi)$ denote the original rendering and $x_\delta = R(\mathcal{G}_s + \delta; \pi)$ the perturbed rendering, then we re-formulate Eq.~\ref{objective} as:
\begin{equation}
    \min_{\delta} \; \mathcal{L}\big(f_\theta(x_\delta), f_\theta(I_{\mathrm{tar}})\big)
\quad \text{s.t.} \quad I_{\mathrm{tar}} \neq x,
\end{equation}
\vspace{-0.1pt}
where $\mathcal{L}$ denotes a similarity measure in the VLM embedding space. The perturbation $\delta$ is optimized such that the embedding of the rendered adversarial image aligns with that of the target image, while remaining visually and geometrically consistent with the original face. This formulation of \methodname~explicitly drives the VLM representation of the perturbed face toward a pose-consistent yet semantically divergent target embedding.

\vspace{-0.2in}

\subsection{Gaussian Shell Initialization}
%\vspace{-0.1in}
For the adversarial noise training, we initialize from a pre-optimized 3DGS representation $\mathcal{G}_s$ consisting of $N_{\text{base}}$ Gaussians while keeping all baseline parameters fixed throughout optimization to preserve geometry and identity.
We instantiate a secondary set of $N_{\text{shell}}$ auxiliary Gaussians, referred to as a \emph{Gaussian shell} to parameterize the additive perturbation $\delta$.
Each Gaussian shell is initialized by randomly sampling the spatial mean of a baseline Gaussian and applying a small uniform perturbation $\epsilon \sim \mathcal{U}([-0.01, 0.01]^3)$, ensuring close spatial proximity to the original surface. The covariance, rotation, and spatial parameters of the shell Gaussians are frozen after initialization. Also, optimization is restricted to their appearance attributes, specifically color and opacity, which together define the additive perturbation $\delta$ in the rendered space. This constrained parameterization enforces that adversarial manipulation operates purely through localized appearance and density modulation, yielding semantically effective yet geometrically non-destructive perturbations of the underlying 3D face.

\subsection{Multi-View Adversarial Noise Training}

%Optimization is performed using a simultaneous multi-view rasterization approach. At each iteration, the augmented 3D model is rendered as a 2D image at a resolution of 384x384 with a $50\degree$ field of view and a fixed camera radius of 2.7. We render from five fixed camera trajectories: frontal ($270\degree$, $0\degree$), slight left ($255\degree$, $0\degree$), slight right ($285\degree$, $0\degree$), slight up ($270\degree$, $15\degree$), and slight down ($270\degree$, $-15\degree$). To evaluate generalizability and prevent overfitting to the optimization angles, we introduce a sixth novel view of ($285\degree$, $15\degree$). This view is explicitly excluded from the optimization objective. This view is exclusively used for validation. 
At each optimization step of \methodname, the perturbed 3D representation $\mathcal{G}_s + \delta$ is rendered into $K$ differentiable views $\{x_k\}_{k=1}^{K}$ using a predefined camera set $\Pi = \{\pi_k\}_{k=1}^{K}$, where each camera $\pi_k$ is parameterized by azimuth $\theta_k$, elevation $\phi_k$, radius $r$, and field of view $\psi$. Formally,
\begin{equation}
    x_k = R(\mathcal{G}_s + \delta; \pi_k)
\end{equation}
The adversarial objective of \methodname~is computed over a fixed training subset of $\Pi$ to enforce cross-view consistency of the perturbation in 3D space.
We apply stochastic augmentations to each rendered view prior to encoding to prevent overfitting to specific viewpoints. Specifically, we apply a set of augmentations, denoted by $\mathcal{A}(\cdot)$, simulate 8-bit image discretization via quantization noise and apply random resized crops and color jittering. 
The augmented rendered views $x_k$ and the synthetic multi-attribute target image $I_{\mathrm{tar}}$ are processed by the visual encoder $\mathrm{VE}(\cdot)$ of the pretrained VLM $f_\theta$ to obtain their corresponding feature embeddings:
$z_k = \mathrm{VE}(\mathcal{A}(x_k))$ and 
$z_{\mathrm{tar}} = \mathrm{VE}(\mathcal{A}(I_{\mathrm{tar}}))$.
Then, we optimize $\delta$ to align these feature embeddings to guide the attribute transfer through image injection. In practice, the perturbation $\delta$ is optimized using features extracted from $E$ different visual encoders, corresponding to the visual backbones of multiple VLMs. This multi-encoder supervision encourages consistent feature alignment across diverse representations, improving robustness and enabling the learned perturbations to generalize in a model-agnostic manner.
Thus, the feature alignment loss aggregates over all training views as
\begin{equation}
\mathcal{L}_{\mathrm{feature}} 
= \frac{1}{EK}\sum_{e=1}^{E}\sum_{k=1}^{K}
\Big(1 - \mathrm{sim}\big(z_k, z_{\mathrm{tar}}\big)\Big),
\label{eq:feature_loss}
\end{equation}
where $\mathrm{sim}(\mathbf{a},\mathbf{b}) = \frac{\mathbf{a}^\top \mathbf{b}}{\|\mathbf{a}\|_2 \|\mathbf{b}\|_2}$ denotes cosine similarity. The overall optimization jointly minimizes $\mathcal{L}_{\mathrm{feature}}$ and regularization terms on the shell parameters, yielding semantically targeted yet geometrically consistent adversarial perturbations across viewpoints.

\vspace{-0.1in}

\subsection{Face-Aware Regularization Mechanisms}

Finally, we introduce face-aware regularization terms to preserve identity and photorealism of the underlying 3D face. While $\mathcal{L}_{\mathrm{feature}}$ drives semantic embedding shifts, the following constraints suppress unnecessary perturbations and ensure that the learned noise remains localized and structurally consistent.

\paragraph{Landmark Consistency.}
In order to protect identity-critical regions, we extract fixed binary masks $M$ corresponding to sensitive facial areas (e.g., \textit{eyes}, \textit{nose}, and \textit{lips}) from the baseline renders. Let $I_k$ and $I^{\mathrm{base}}_k$ denote the current and baseline renders at view $\pi_k$, respectively. We penalize pixel-level deviations within the masked regions across all training views:
\begin{equation}
\mathcal{L}_{\mathrm{landmarks}} 
=  \frac{1}{K} \sum_{k=1}^{K}
\left\| M \odot (I_k - I^{\mathrm{base}}_k) \right\|_2^2,
\label{eq:landmark_loss}
\end{equation}
where $\odot$ denotes element-wise multiplication. This constraint discourages distortions in identity-sensitive areas while allowing controlled modifications elsewhere.

% \paragraph{Chroma-Constrained Texture Regularization.}
% Texture-based adversarial perturbations often manifest as high-frequency luminance artifacts. To mitigate this effect, we map rendered images to the YUV color space and regularize perturbations preferentially in the luminance channel. Let $(Y_k, U_k, V_k)$ denote the YUV decomposition of $I_k - I^{\mathrm{base}}_k$. We define:
% \begin{equation}
% \mathcal{L}_{\mathrm{chroma}} 
% = \lambda_{\mathrm{visual}} 
% \frac{1}{K} \sum_{k=1}^{K}
% \left( \lambda_Y \|Y_k\|_2^2 + \|U_k\|_2^2 + \|V_k\|_2^2 \right),
% \label{eq:chroma_loss}
% \end{equation}
% where $\lambda_Y$ controls the strength of luminance suppression. To promote stable optimization, the visual regularization weight is gradually annealed using a cosine schedule. Let $p = \frac{i}{T}$ denote the normalized training progress at iteration $i$ out of $T$ total iterations, and define the cosine progression factor $C = \frac{1}{2}(1 - \cos(\pi p))$. The iteration-dependent weight is then given by $\lambda_{\mathrm{visual}} = \lambda_0 C$, where $\lambda_0$ is a constant scaling factor.

\paragraph{Chroma-Constrained Texture Regularization.}
Texture-based adversarial perturbations often manifest as high-frequency luminance artifacts. We mitigate this effect and encourage smooth semantic transfer by mapping the additive texture parameters of the auxiliary shell Gaussians directly into the YUV color space, emphasizing our regularization in the luminance channel. Let $(Y_i, U_i, V_i)$ denote the YUV decomposition of the explicit texture noise applied to the $i$-th shell Gaussian, out of $N_{\textrm{shell}}$ total auxiliary Gaussians. We define:
\begin{equation}
\mathcal{L}_{\mathrm{chroma}} 
= \frac{1}{N_{\text{shell}}} \sum_{i=1}^{N_{\text{shell}}} \left( \lambda_{Y} Y_i^2 + U_i^2 + V_i^2 \right),
\label{eq:chroma_loss}
\end{equation}
\vspace{-0.3in}
%
%where $\lambda_Y$ controls the strength of the luminance supression. 

\subsection{Overall Optimization of \methodname}

\begin{figure*}[h]
    \centering
    \scalebox{0.8}{
    \includegraphics[width=1\linewidth]{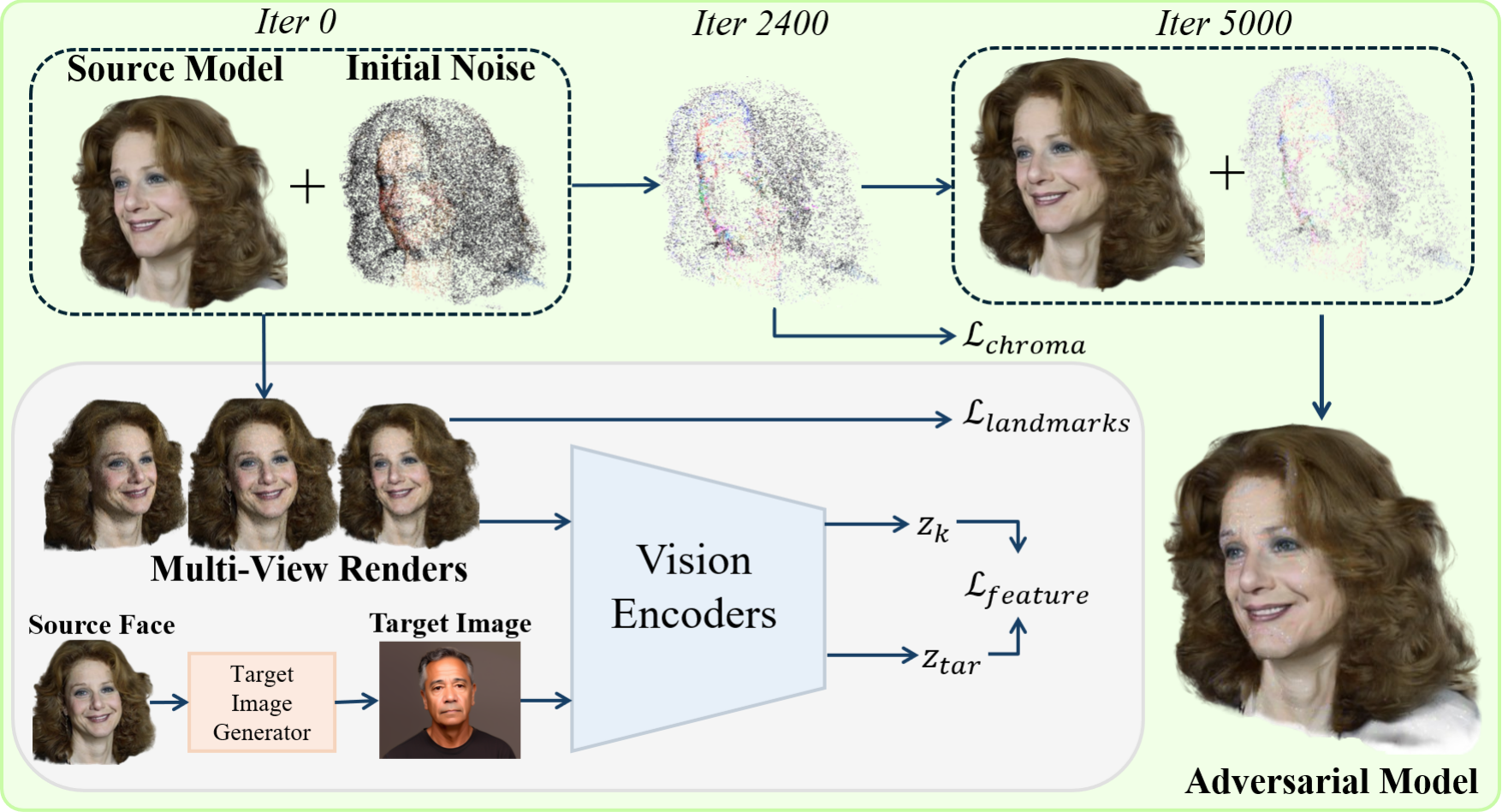}}
    \caption{The 3D FaceShell Optimization Pipeline. \textbf{3D Operations (Green):} A learnable Gaussian shell is initialized over a frozen 3DGS baseline and shell parameters are iteratively optimized and Gaussian color is regulated ($\mathcal{L}_{\mathrm{chroma}}$). \textbf{2D Operations (Gray):} Rendered views are generated to guide feature alignment in shell Gaussians ($\mathcal{L}_{\mathrm{feature}}$) and enforce identity preservation ($\mathcal{L}_{\mathrm{landmarks}}$).} \vspace{-0.1in}
    \label{Method}
\end{figure*}

The optimization of \methodname\ jointly balances semantic embedding manipulation and face-aware regularization in the 3D Gaussian shell parameter space as illustrated in Fig.~\ref{Method}. At each iteration, gradients from the multi-view feature alignment loss are backpropagated through the differentiable renderer to update only the shell parameters (color and opacity), while the baseline Gaussians remain frozen. The overall objective combines latent feature matching with structural and photometric constraints:
\begin{equation}
\mathcal{L}_{\mathrm{total}} 
= \mathcal{L}_{\mathrm{feature}} 
+ \lambda_{\mathrm{chroma}} \mathcal{L}_{\mathrm{chroma}} 
+ \lambda_{\mathrm{landmarks}}\mathcal{L}_{\mathrm{landmarks}}.
\label{eq:total_loss}
\end{equation}
where $\lambda_{\mathrm{chroma}}$ and $\lambda_{\mathrm{landmarks}}$ are the strengths of landmark consistency and luminance suppression, respectively.
Together, these losses constrain the adversarial optimization to produce semantically effective yet visually coherent perturbations, preserving facial identity and structural realism across viewpoints.

\vspace{-1.25em}
\section{Experiments}

\subsection{Implementation Details}
%Write both the progression schedules.
%Mention the encoder E used.

%To promote stable optimization, the visual regularization weight is dynamically annealed using a cosine progression schedule. Let $p = \frac{i}{T}$ denote the normalized training progress at iteration $i$ out of $T$ total iterations, and define the cosine progression factor $C = \frac{1}{2}(1 - \cos(\pi p))$. The iteration-dependent weight is then given by $\lambda_{\mathrm{visual}} = \lambda_0 C$, where $\lambda_0$ is a constant scaling factor.

%To further regularize the optimization and prevent opacity-driven geometric artifacts, we impose a progressively tightening upper bound on the shell Gaussian opacity. The maximum allowable opacity is dynamically clamped as
%\[
%O_{\max} = -0.5 - 2.5C.
%\]
%where $C$ is the cosine progression factor.
%As training progresses, this schedule gradually pushes shell Gaussians toward increased transparency, suppressing perturbations that do not meaningfully contribute to the semantic objective. This progressive clamping acts as an additional geometric regularizer, ensuring that the adversarial signal remains minimally intrusive while preserving structural fidelity of the underlying 3D face.

For the adversarial feature alignment in \methodname, we use a multi-encoder training with $E=2$ distinct vision encoders: SigLIP (ViT-SO400M-Patch14)\cite{siglip} and CLIP (ViT-Large-Patch14-336)\cite{clip}. %Aggregating the cosine similarity losses across multiple representation spaces prevents the Gaussian shell from overfitting to a single model's feature space, ultimately yielding more robust adversarial perturbations.
For stable optimization, the weight of the chroma-constrained texture regularization is dynamically annealed using a cosine progression schedule. If $p = \frac{i}{T}$ denote the normalized training progress at iteration $i$ out of $T$ total iterations, and define the cosine progression factor $C = \frac{1}{2}(1 - \cos(\pi p))$. The iteration-dependent weight is then given by $\lambda_{\mathrm{chroma}} = \lambda_0 C$, where $\lambda_0$ is a constant scaling factor. Additionally, we further regularize the optimization and prevent opacity-driven geometric artifacts by imposing a progressively tightening upper bound on the shell Gaussian opacity. The maximum allowable opacity is dynamically clamped as
$
O_{\max} = -0.5 - 2.5C.
$
As training progresses, this schedule gradually pushes Gaussian shells toward increased transparency, suppressing perturbations that do not meaningfully contribute to the semantic objective. This progressive clamping acts as an additional geometric regularizer, ensuring that the adversarial signal remains minimally intrusive while preserving structural fidelity of the underlying 3D face.

\textbf{Hyperparameters.} 
%Mention $N_{shell}=50,000$
% $\lambda_{\mathrm{chroma}}$ and $\lambda_{\mathrm{landmarks}}$
Unless otherwise specified, rendered images are generated at a resolution of $384 \times 384$ with a field of view $\psi = 50^\circ$ and a fixed camera radius $r = 2.7$. We use $K=5$ fixed camera poses for optimization: frontal $(270^\circ, 0^\circ)$, slight left $(255^\circ, 0^\circ)$, slight right $(285^\circ, 0^\circ)$, slight up $(270^\circ, 15^\circ)$, and slight down $(270^\circ, -15^\circ)$. To evaluate cross-view generalization, we introduce a novel validation view at $(285^\circ, 15^\circ)$, which is excluded from optimization and used exclusively for evaluation. We optimize the color and opacity parameters of $N_{shell}=50,000$ shell Gaussians using the Adam optimizer for 10,000 total iterations, setting the learning rate to 0.1 for both parameter groups. For our face-aware regularization mechanisms, the landmark consistency loss weight is fixed at $\lambda_{\mathrm{landmarks}}=500$. In the chroma-constrained texture regularization, the luminance penalty is enforced with $\lambda_Y=10.0$ while we scale our overall chroma-constrained texture loss with $\lambda_{\mathrm{chroma}}=0.4$. Finally, during the multi-view adversarial noise training, the stochastic augmentations $\mathcal{A}(\cdot)$ applied to the rendered views consist of simulated 8-bit quantization noise (uniformly sampled between ± 2/255), a random resized crop with a scale factor between 0.8 and 1.0, and random color jittering adjusting brightness and contrast by up to 10\%.

\vspace{-1.2em}
\subsection{Evaluation Settings}
\textbf{Dataset.} In order to evaluate our method and the competing baselines, we construct a benchmark consisting of 100 celebrity faces curated from the IMDB-WIKI dataset~\cite{IMDB}. For each subject, the facial attributes \textit{eye color}, \textit{hair color}, \textit{hair type}, and \textit{expression} are manually annotated, while \textit{age}, \textit{gender}, and \textit{ethnicity} are obtained from publicly available metadata corresponding to the time the source photograph was captured. Each 2D face image is then converted into a 3D Gaussian Splatting (3DGS) representation using FaceLift~\cite{facelift}. The resulting benchmark, comprising 3D faces paired with curated attribute annotations, will be publicly released to facilitate further research in this direction.

\noindent  \textbf{Metrics.} We assess the capabilities of each attack method to successfully transfer target attributes onto rendered views of each source 3D face avatar. We provide the rate at which a given VLM outputs target attributes when provided with a rendered view of the source face (Injection Rate) and the rate at which a given VLM outputs an attribute different from the ground truth attribute of a given source face when provided with a rendered view of that source face (Mismatch Rate). Importantly, we assess the quality of the outputs for each method using LPIPS~\cite{LPIPS} to measure the perceptual distance between the synthesized images and the ground truth references. Finally, we utilize ArcFace \cite{ArcFace} to determine the identity similarity between the generated face views produced by each method and the faces in the ground truth images. 

\begin{wrapfigure}{r}{0.48\textwidth}
\vspace{-0.4in}
\centering
\includegraphics[width=\linewidth]{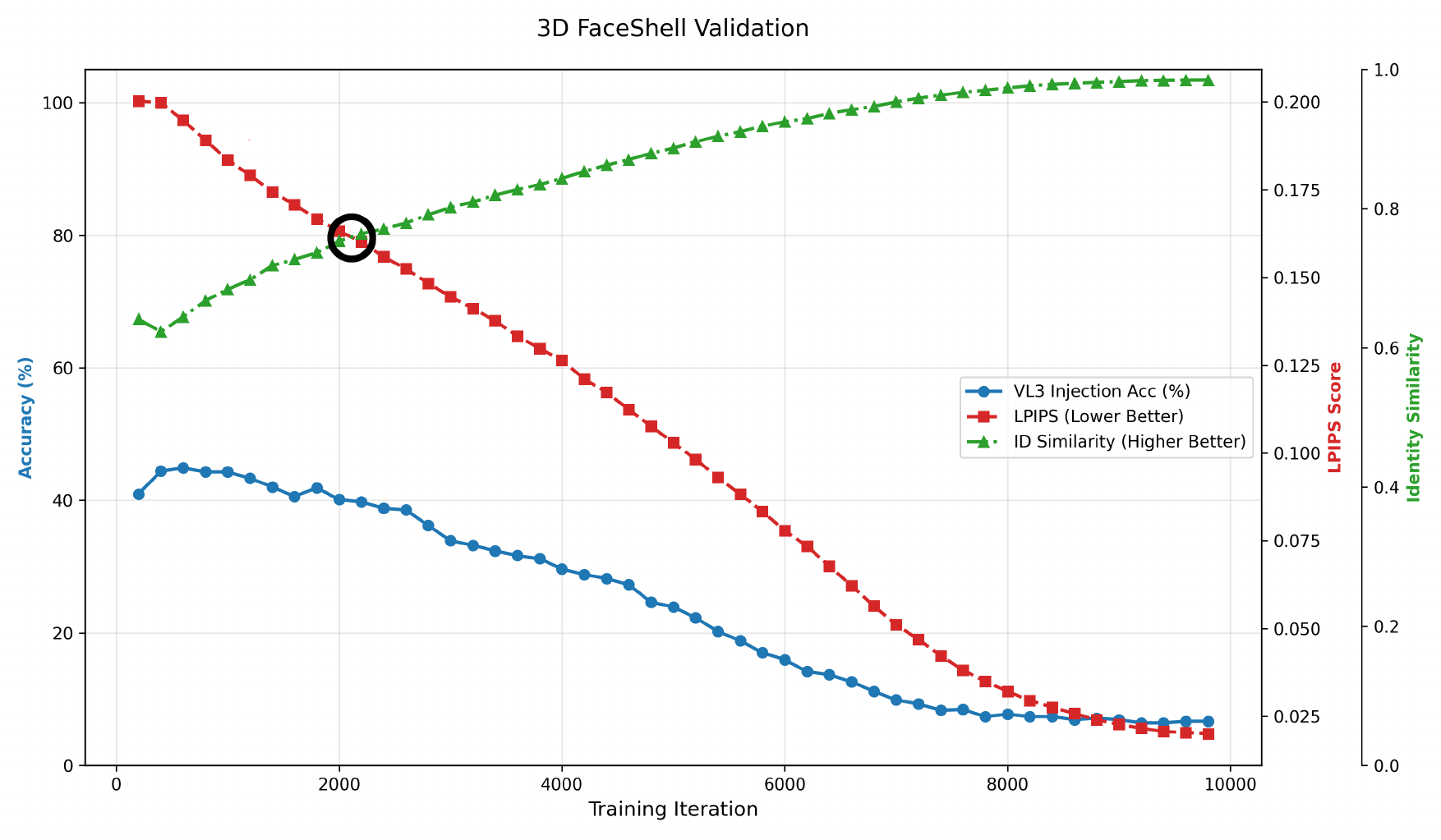}
\caption{Validation results on VideoLLaMA3 used to select the best-performing samples during optimization.}
\vspace{-0.2in}
\label{fig:validation}
\end{wrapfigure}
\noindent \textbf{Selection Criteria.} To determine the optimal stopping point for our adversarial optimization and ensure consistent perceptual fidelity, we evaluate the training progression on a validation set consisting of 20 separate faces.  As illustrated in Fig. \ref{fig:validation}, the optimization process reveals a clear trade-off between the VLM injection accuracy and the perceptual quality (LPIPS) over the course of 10,000 training iterations. To ensure our method balances high visual fidelity with maximum semantic manipulation, we apply a strict threshold-based selection criterion for our final evaluations. For each source face, we extract the rendering from the iteration that yields the highest injection accuracy, strictly bottlenecked by the requirement that the LPIPS score must remain below 0.165. In edge cases where the optimization fails to produce any samples under this 0.165 threshold, we default to selecting the iteration that achieved the absolute lowest LPIPS score overall. This guarantees that the final perturbed 3D avatars remain perceptually grounded and structurally consistent, preventing unbounded visual degradation while still capturing the peak adversarial transfer rate.

\noindent \textbf{Models.} We evaluate all methods on four black-box VLMs: VideoLLaMA3~\cite{VL3}, LLaVA-NeXT~\cite{llavanext}, LLaVA-OV~\cite{llavaone}, and BLIP-2~\cite{blip2}. These models are selected due to their strong multimodal reasoning capabilities and the diversity of their visual encoders. In particular, BLIP-2 employs a CLIP-based visual encoder, whereas the others utilize SigLIP-based visual encoders. %This diversity enables us to evaluate the robustness and transferability of our method across VLMs with different visual backbone architectures.

\noindent  \textbf{Baselines.} As there are no other methods for adversarial semantic editing in the 3DGS space, we benchmark \methodname~against six state-of-the-art 2D adversarial attacks targeting VLMs: AnyAttack~\cite{anyattack}, V-Attack Ensemble~\cite{vattack}, AdvDiffVLM~\cite{advdiff}, SSA-CWA~\cite{ssacwa}, M-Attack~\cite{mattack}, and AttackVLM~\cite{attackvlm}. We selected these specific baselines because they represent a comprehensive cross-section of the most effective modern 2D attack paradigms, including diffusion-based, self-supervised, and highly transferable embedding-space perturbations.

\vspace{-0.1in}

\subsection{Quantitative Results}

\textbf{State of the Art Comparison.} As shown in Table~\ref{tab:main_results}, \methodname\ achieves competitive performance in terms of both Injection Rate (IR) and Mismatch Rate (MR) across all evaluated vision-language models. While aggressive 2D baselines such as M-Attack and V-Attack obtain slightly higher mismatch rates overall and higher injection rates on certain models (e.g., BLIP-2 and LLaVA-NeXT), the semantic performance gap remains small. For instance, \methodname\ trails the strongest baseline (M-Attack) by only a 3.9\% relative difference in LLaVA-OV MR and by 10.3\% relative difference compared to V-Attack on VideoLLaMA3 MR. Moreover, \methodname\ surpasses both M-Attack and V-Attack in absolute Injection Rate on VideoLLaMA3 and LLaVA-OV.
Importantly, the marginal semantic gains of these aggressive 2D approaches come at the cost of significant degradation in \textit{visual quality} and \textit{identity preservation}. In contrast, \methodname\ maintains substantially higher perceptual and structural fidelity, achieving a \textcolor{ForestGreen}{+71.5\%} improvement in LPIPS compared to V-Attack and a \textcolor{ForestGreen}{+60.7\%} improvement over M-Attack. Similarly, \methodname\ improves Identity Similarity by \textcolor{ForestGreen}{+88.9\%} and \textcolor{ForestGreen}{+56.2\%} relative to V-Attack and M-Attack, respectively. This distinction is critical, as preserving facial identity and geometric structure is essential for realistic and minimally intrusive adversarial manipulation.

\noindent \textbf{Multi-Attribute vs Single-Attribute Transfer.} Table \ref{tab:attribute_results} presents a quantitative comparison of Transfer Rate and Mismatch Rate on VideoLLaMA3 when modifying all targeted semantic attributes simultaneously versus isolating a single attribute and keeping the rest constant. Our results indicate that shifting all attributes at once yields superior performance for most individual attributes. For instance, transferring all attributes simultaneously results in a \textcolor{ForestGreen}{+50.7\%} relative increase in the transfer rate for hair color and a \textcolor{ForestGreen}{+96.7\%} relative increase in the transfer rate for expression compared to isolated single-attribute attacks. Additionally, Gender and Ethnicity transfer rates see relative increases of \textcolor{ForestGreen}{+3.2\%} and \textcolor{ForestGreen}{+9.7\%}, respectively. This result confirms that inducing a compounded semantic shift via multi-attribute target image injection provides a stronger gradient signal, thereby improving the attack success rate across individual attributes.
\vspace{-0.8pt}

\begin{table}[h]
\centering
\caption{Quantitative comparison of Injection Rate (IR) and Mismatch Rate (MR) across multiple vision-language models, alongside perceptual (LPIPS) and structural (Identity Similarity) preservation metrics across frontal face views.}
\label{tab:main_results}
\resizebox{\textwidth}{!}{
\begin{tabular}{l cc cc cc cc >{\columncolor{blue!6}}c >{\columncolor{blue!6}}c}
\toprule
\multirow{2}{*}{\textbf{Method}} 
& \multicolumn{2}{c}{\textbf{VideoLLaMA3 (\%)}} 
& \multicolumn{2}{c}{\textbf{LLaVA-NeXT (\%)}} 
& \multicolumn{2}{c}{\textbf{LLaVA-OV (\%)}} 
& \multicolumn{2}{c}{\textbf{BLIP-2 (\%)}} 
& \multirow{2}{*}{\textbf{LPIPS $\downarrow$}} 
& \multirow{2}{*}{\textbf{ID Sim $\uparrow$}} \\

\cmidrule(lr){2-3} \cmidrule(lr){4-5} \cmidrule(lr){6-7} \cmidrule(lr){8-9}

& \textbf{IR} & \textbf{MR} 
& \textbf{IR} & \textbf{MR} 
& \textbf{IR} & \textbf{MR} 
& \textbf{IR} & \textbf{MR} 
& & \\

\midrule

Nothing             
& 5.9  & 19.3  
& 4.6  & 17.1  
& 4.4  & 15.7  
& 13.1 & 39.7  
& \cellcolor{White!}0.0000 & \cellcolor{White!}1.0000 \\

Random Noise        
& 8.1  & 22.5  
& 6.9  & 21.6  
& 6.9  & 20.6  
& 12.5 & 39.2  
& \cellcolor{White!}0.0862 & \cellcolor{White!}0.9535 \\

\midrule

AnyAttack\cite{anyattack}            
& \cellcolor{ForestGreen!10} 6.6  & \cellcolor{ForestGreen!10} 22.9  
& \cellcolor{ForestGreen!10} 5.7  & \cellcolor{ForestGreen!10} 21.1  
& \cellcolor{ForestGreen!10} 6.4  & \cellcolor{ForestGreen!10} 19.3  
& \cellcolor{ForestGreen!13} 13.6 & \cellcolor{ForestGreen!14} 41.1  
& \cellcolor{ForestGreen!12} 0.4452 & \cellcolor{ForestGreen!42} \underline{0.7795} \\

AttackVLM\cite{attackvlm}            
& \cellcolor{ForestGreen!12} 8.9  & \cellcolor{ForestGreen!11} 24.7  
& \cellcolor{ForestGreen!11} 7.1  & \cellcolor{ForestGreen!10} 21.0  
& \cellcolor{ForestGreen!11} 7.6  & \cellcolor{ForestGreen!11} 21.3  
& \cellcolor{ForestGreen!10} 11.4 & \cellcolor{ForestGreen!10} 38.3  
& \cellcolor{ForestGreen!38} \underline{0.2428} & \cellcolor{ForestGreen!50} \textbf{0.8506} \\

AdvDiffVLM\cite{advdiff}          
& \cellcolor{ForestGreen!10} 6.9  & \cellcolor{ForestGreen!11} 23.6  
& \cellcolor{ForestGreen!11} 6.7  & \cellcolor{ForestGreen!11} 22.1  
& \cellcolor{ForestGreen!10} 5.9  & \cellcolor{ForestGreen!12} 21.9  
& \cellcolor{ForestGreen!11} 12.1 & \cellcolor{ForestGreen!13} 40.6  
& \cellcolor{ForestGreen!35} 0.2638 & \cellcolor{ForestGreen!28} 0.6526 \\

SSA-CWA\cite{ssacwa}             
& \cellcolor{ForestGreen!36} 33.9 & \cellcolor{ForestGreen!37} 57.0  
& \cellcolor{ForestGreen!26} 21.7 & \cellcolor{ForestGreen!27} 41.6  
& \cellcolor{ForestGreen!30} 32.0 & \cellcolor{ForestGreen!31} 50.4  
& \cellcolor{ForestGreen!50} \underline{37.7} & \cellcolor{ForestGreen!48} \underline{68.3}  
& \cellcolor{ForestGreen!10} 0.4626 & \cellcolor{ForestGreen!17} 0.5513 \\

V-Attack\cite{vattack}            
& \cellcolor{ForestGreen!46} 44.4 & \cellcolor{ForestGreen!50} \textbf{74.0}  
& \cellcolor{ForestGreen!40} \underline{35.7} & \cellcolor{ForestGreen!44} \underline{63.4}  
& \cellcolor{ForestGreen!46} 51.6 & \cellcolor{ForestGreen!48} \underline{75.1}  
& \cellcolor{ForestGreen!44} 33.9 & \cellcolor{ForestGreen!45} 66.6  
& \cellcolor{ForestGreen!18} 0.4039 & \cellcolor{ForestGreen!14} 0.5255 \\

M-Attack\cite{mattack}            
& \cellcolor{ForestGreen!49} \underline{47.4} & \cellcolor{ForestGreen!47} \underline{69.6}  
& \cellcolor{ForestGreen!50} \textbf{46.3} & \cellcolor{ForestGreen!50} \textbf{70.3}  
& \cellcolor{ForestGreen!50} \underline{56.4} & \cellcolor{ForestGreen!50} \textbf{77.4}  
& \cellcolor{ForestGreen!50} \textbf{37.9} & \cellcolor{ForestGreen!50} \textbf{70.3}  
& \cellcolor{ForestGreen!20} 0.3814 & \cellcolor{ForestGreen!10} 0.4884 \\

\midrule
\cellcolor{ForestGreen!15} \textbf{\methodname}        
& \cellcolor{ForestGreen!50} \textbf{48.7} & \cellcolor{ForestGreen!44} 66.4  
& \cellcolor{ForestGreen!35} 31.3 & \cellcolor{ForestGreen!36} 53.0  
& \cellcolor{ForestGreen!50} \textbf{56.9} & \cellcolor{ForestGreen!48} 74.4  
& \cellcolor{ForestGreen!33} 26.7 & \cellcolor{ForestGreen!29} 53.7  
& \cellcolor{ForestGreen!50} \textbf{0.1499} & \cellcolor{ForestGreen!40} 0.7629 \\

\bottomrule
\end{tabular}\vspace{-0.2in}
}
\end{table}

\vspace{-1.2em}
\begin{table*}
\centering
\caption{Quantitative comparison of Transfer Rate (TR) and Mismatch Rate (MR) across training and validation views for each targeted semantic attribute on VideoLLaMA3.} \vspace{-0.1in}
\label{tab:attribute_results}
\resizebox{\textwidth}{!}{
\begin{tabular}{l cc cc cc cc cc cc cc}
\toprule
\multirow{2}{*}{\textbf{Method}} 
& \multicolumn{2}{c}{\textbf{Age (\%)}} 
& \multicolumn{2}{c}{\textbf{Gender (\%)}} 
& \multicolumn{2}{c}{\textbf{Ethnicity (\%)}} 
& \multicolumn{2}{c}{\textbf{Expression (\%)}} 
& \multicolumn{2}{c}{\textbf{Hair (\%)}} 
& \multicolumn{2}{c}{\textbf{Hair Color (\%)}} 
& \multicolumn{2}{c}{\textbf{Eye Color (\%)}} \\

\cmidrule(lr){2-3} \cmidrule(lr){4-5} \cmidrule(lr){6-7} \cmidrule(lr){8-9} \cmidrule(lr){10-11} \cmidrule(lr){12-13} \cmidrule(lr){14-15}

& \textbf{IR} & \textbf{MR} 
& \textbf{IR} & \textbf{MR} 
& \textbf{IR} & \textbf{MR} 
& \textbf{IR} & \textbf{MR} 
& \textbf{IR} & \textbf{MR} 
& \textbf{IR} & \textbf{MR} 
& \textbf{IR} & \textbf{MR} \\

\midrule

\textbf{\methodname}
& 40.8 & 83.3 
& 64.1 & 64.1 
& 34.0 & 68.3 
& 17.9 & 49.7 
& 37.6 & 60.2 
& 21.1 & 72.5 
& 50.4 & 70.3 \\

Individual
& 35.6 & 82.0 
& 62.1 & 62.1 
& 31.0 & 69.1 
& 9.1  & 50.0 
& 31.8 & 63.1 
& 14.0 & 58.7 
& 57.3 & 42.7 \\

\bottomrule
\end{tabular}
}
\end{table*} \vspace{-0.5in}

\iffalse
\begin{table*}[t]
\centering
\caption{Quantitative comparison of training view results and novel validation only view results.}\vspace{-0.1in}
\label{tab:view_results}
\resizebox{\textwidth}{!}{
\begin{tabular}{l cc cc cc cc cc}
\toprule
\multirow{2}{*}{Method} & \multicolumn{2}{c}{VideoLLaMA3 (\%)} & \multicolumn{2}{c}{LLaVA-NeXT (\%)} & \multicolumn{2}{c}{LLaVA-OV (\%)} & \multicolumn{2}{c}{BLIP-2 (\%)} & \multirow{2}{*}{LPIPS $\downarrow$} & \multirow{2}{*}{ID Sim $\uparrow$} \\
\cmidrule(lr){2-3} \cmidrule(lr){4-5} \cmidrule(lr){6-7} \cmidrule(lr){8-9}
& Inj & Mis & Inj & Mis & Inj & Mis & Inj & Mis & & \\
\midrule
Ours - Training Views           & \textbf{46.8} & \textbf{65.2} & \textbf{30.3} & \textbf{52.8} & \textbf{56.0} & \textbf{73.5} & \textbf{26.2} & \textbf{53.2} & 0.1523 & \underline{0.7550} \\
Ours - Validation View            & \underline{36.4} & \underline{56.3} & \underline{23.6} & \underline{45.0} & \underline{41.1} & \underline{59.1} & \underline{23.3} & \underline{51.1} & \textbf{0.1465} & 0.7289 \\
Front View Training - Validation View            & 23.4 & 42.0 & 13.7 & 33.0 & 26.6 & 44.1 & 17.1 & 45.3 & \underline{0.1474} & \textbf{0.7596} \\
\bottomrule
\end{tabular}
}
\end{table*}
\fi 
\begin{table*}
\centering
\caption{Quantitative comparison between training-view results and novel validation-view results.}
\label{tab:view_results}
\resizebox{\textwidth}{!}{
\begin{tabular}{l cc cc cc cc >{\columncolor{blue!6}}c >{\columncolor{blue!6}}c}
\toprule

\multirow{2}{*}{\textbf{Method}} 
& \multicolumn{2}{c}{\textbf{VideoLLaMA3 (\%)}} 
& \multicolumn{2}{c}{\textbf{LLaVA-NeXT (\%)}} 
& \multicolumn{2}{c}{\textbf{LLaVA-OV (\%)}} 
& \multicolumn{2}{c}{\textbf{BLIP-2 (\%)}} 
& \multirow{2}{*}{\textbf{LPIPS $\downarrow$}} 
& \multirow{2}{*}{\textbf{ID Sim $\uparrow$}} \\

\cmidrule(lr){2-3}
\cmidrule(lr){4-5}
\cmidrule(lr){6-7}
\cmidrule(lr){8-9}

& \textbf{IR} & \textbf{MR}
& \textbf{IR} & \textbf{MR}
& \textbf{IR} & \textbf{MR}
& \textbf{IR} & \textbf{MR}
& & \\

\midrule

\rowcolor{ForestGreen!6}
\textbf{Ours -- Training Views}
& \textbf{46.8} & \textbf{65.2}
& \textbf{30.3} & \textbf{52.8}
& \textbf{56.0} & \textbf{73.5}
& \textbf{26.2} & \textbf{53.2}
& 0.1523 & \underline{0.7550} \\

\textbf{Ours -- Validation View}
& \underline{36.4} & \underline{56.3}
& \underline{23.6} & \underline{45.0}
& \underline{41.1} & \underline{59.1}
& \underline{23.3} & \underline{51.1}
& \textbf{0.1465} & 0.7289 \\

Front View Training -- Validation View
& 23.4 & 42.0
& 13.7 & 33.0
& 26.6 & 44.1
& 17.1 & 45.3
& \underline{0.1474} & \textbf{0.7596} \\

\bottomrule
\end{tabular}
}
\end{table*} \vspace{-0.2in}

\textbf{Commercial Model Evaluation.} Table \ref{tab:gemini_results} extends our primary evaluation to the commercial Gemini-2.5-Flash\cite{gemini} VLM, comparing our method against our 2D adversarial attack baselines. Our \methodname~successfully generalizes to this closed-source model, achieving a competitive injection rate and mismatch rates. Crucially, while our method observes a relatively low difference in semantic manipulation metrics, falling short of V-Attack's injection rate by a relative 13.4\% and M-Attack's by a relative 28.3\%, it yields a disproportionately massive improvement in visual fidelity. Specifically, 3D FaceShell achieves an LPIPS score of 0.1499, representing an impressive 62.9\% relative improvement over V-Attack and a 60.7\% improvement over M-Attack. Similarly, \methodname~achieves an identity similarity score of 0.7629, structurally outperforming V-Attack\cite{vattack} by 45.2\% and M-Attack\cite{mattack} by 56.2\%. This demonstrates that \methodname maintains semantic competitiveness without relying on the severe, unrealistic face degradation characteristic of 2D baselines. It should be noted that we additionally tried to evaluate each of the methods on GPT-4o\cite{gpt} but this model largely refused to answer questions about face attributes despite prompt engineering efforts.

\begin{wraptable}{r}{0.48\textwidth}
\vspace{-0.2in}
\centering
\caption{Quantitative comparison of Injection Rate (IR) and Mismatch Rate (MR) on Gemini, along with perceptual (LPIPS) and identity preservation (ID Sim) metrics across frontal face views.}
\label{tab:gemini_results}
\resizebox{\linewidth}{!}{
\begin{tabular}{l cc >{\columncolor{blue!6}}c >{\columncolor{blue!6}}c}
\toprule

\multirow{2}{*}{\textbf{Method}} 
& \multicolumn{2}{c}{\textbf{Gemini (\%)}} 
& \multirow{2}{*}{\textbf{LPIPS $\downarrow$}} 
& \multirow{2}{*}{\textbf{ID Sim $\uparrow$}} \\

\cmidrule(lr){2-3}

& \textbf{IR} & \textbf{MR} & & \\

\midrule

Nothing             
& 2.9  & 15.4  
& 0.0000 & 1.0000 \\

Random Noise        
& 4.4  & 19.3  
& 0.0862 & 0.9535 \\

\midrule

AnyAttack\cite{anyattack}            
& 4.0  & 20.7  
& 0.4452 & \underline{0.7795} \\

AttackVLM\cite{attackvlm}            
& 5.4  & 23.6  
& \underline{0.2428} & \textbf{0.8506} \\

AdvDiffVLM\cite{advdiff}          
& 3.9  & 21.7  
& 0.2638 & 0.6526 \\

SSA-CWA\cite{ssacwa}             
& 26.0 & 48.3  
& 0.4626 & 0.5513 \\

V-Attack\cite{vattack}            
& \underline{35.9} & \textbf{76.0}  
& 0.4039 & 0.5255 \\

M-Attack\cite{mattack}            
& \textbf{43.4} & \underline{67.3}  
& 0.3814 & 0.4884 \\

\midrule

\rowcolor{green!6}
\textbf{\methodname}        
& 31.1 & 50.4  
& \textbf{0.1499} & 0.7629 \\

\bottomrule
\end{tabular}
}
\vspace{-0.15in}
\end{wraptable}
\textbf{Cross-View Generalization.} We assess the necessity of our multi-view adversarial noise training in Table \ref{tab:view_results}. When comparing our multi-view training strategy against a model trained exclusively on a single frontal view, it is evident that multi-view training contributes heavily to successful transfer onto novel views. The model trained only on the front view experiences a severe drop in performance on the validation view, whereas our multi-view approach maintains robust generalization. Specifically, multi-view  adversarial noise training achieves a \textcolor{ForestGreen}{+55.6\%} higher relative Injection Rate on the unseen validation view for VideoLLaMA3 compared to the single-view baseline.
\vspace{-0.1in}
\subsection{Qualitative Results}

\begin{figure*}[h]\vspace{-0.2in}
    \centering
    \scalebox{0.9}{
    \includegraphics[width=1\linewidth]{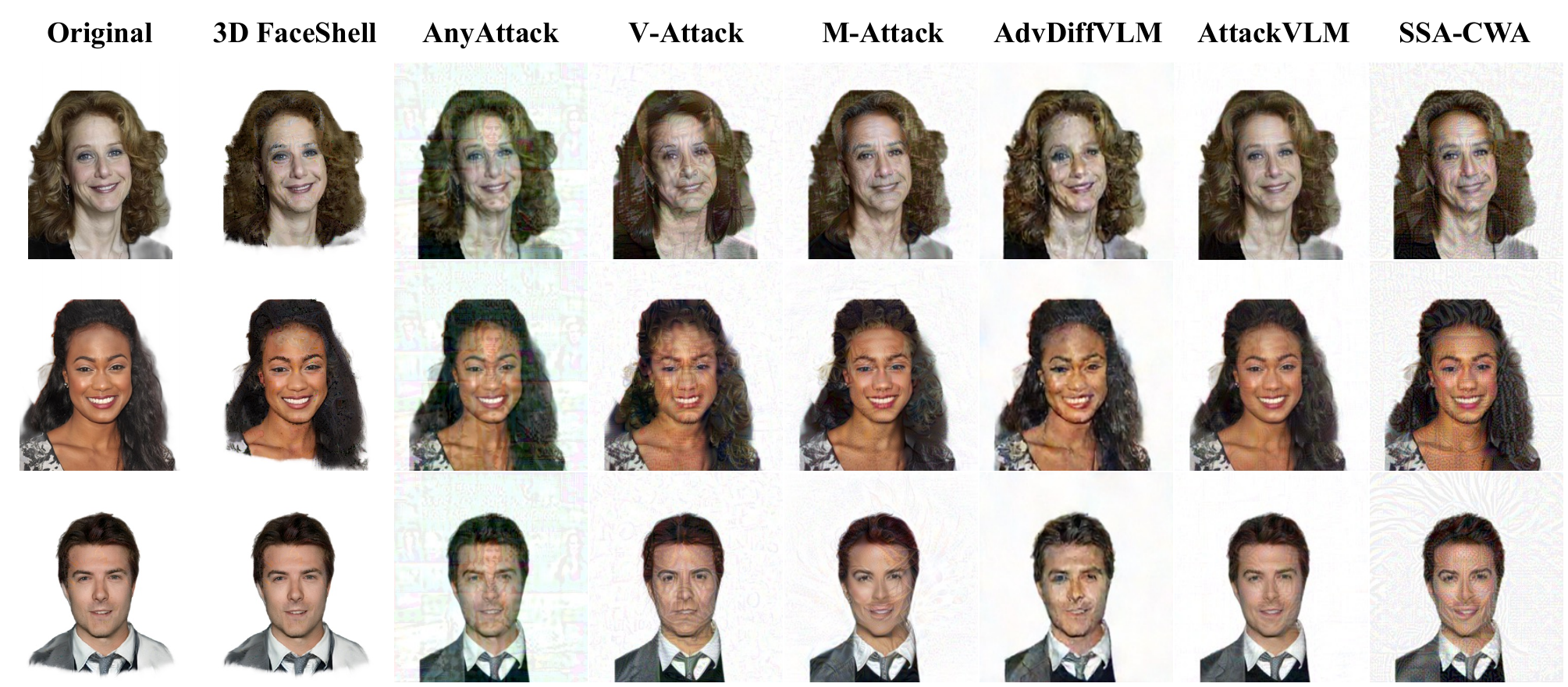}}
    \caption{3D FaceShell achieves superior perceptual fidelity over baselines. While state-of-the-art 2D attacks introduce severe pixelated noise, color distortion, and structural warping to shift VLM perception, our method produces visually inconspicuous perturbations that successfully preserve original photorealism and facial identity.} \vspace{-0.0in}
    \label{Qual}
\end{figure*}

Fig. \ref{Qual} displays the large improvement in identity preservation and visual clarity of our method compared to state-of-the-art 2D methods of semantic shifting. While baseline 2D attacks introduce noticeable pixelated noise, severe color distortions, or structural warping that immediately compromise the photorealism of the face, \methodname~introduces far less intrusive noise. By confining the optimization to our learnable Gaussian shell and heavily regularizing the perturbations, our approach ensures the rendered outputs look naturally persistent. The qualitative comparisons confirm that \methodname~successfully balances the dual objectives of manipulating VLM semantic interpretations while leaving the underlying human-recognizable geometry and textures perceptually undisturbed.

\iffalse
\begin{table*}
\centering
\caption{Ablation study of loss functions and augmentation across all training and validation views.}
\label{tab:loss_results}
\resizebox{\textwidth}{!}{
\begin{tabular}{l cc cc cc cc cc}
\toprule
\multirow{2}{*}{Method} & \multicolumn{2}{c}{VideoLLaMA3 (\%)} & \multicolumn{2}{c}{LLaVA-NeXT (\%)} & \multicolumn{2}{c}{LLaVA-OV (\%)} & \multicolumn{2}{c}{BLIP-2 (\%)} & \multirow{2}{*}{LPIPS $\downarrow$} & \multirow{2}{*}{ID Sim $\uparrow$} \\
\cmidrule(lr){2-3} \cmidrule(lr){4-5} \cmidrule(lr){6-7} \cmidrule(lr){8-9}
& Inj & Mis & Inj & Mis & Inj & Mis & Inj & Mis & & \\
\midrule
No Augmentation           & 17.6 & 34.7 & 16.1 & 33.9 & 28.8 & 45.3 & 15.6 & 42.8 & \textbf{0.1438} & \textbf{0.8397} \\\hline
No Chroma Constrain          & \underline{53.4} & \textbf{72.5} & \textbf{43.1} & \textbf{66.7} & \textbf{61.7} & \textbf{80.4} & \textbf{34.5} & \textbf{61.9} & 0.1911 & 0.5695 \\
No Landmark Consistency          & \textbf{54.6} & \underline{72.2} & \textbf{43.1} & \underline{66.0} & \underline{59.7} & \underline{77.5} & \underline{28.8} & \underline{56.1} & \underline{0.1512} & 0.6184 \\ \hline
No Clamp  & 44.6 & 65.0 & 31.5 & 53.5 & 57.0 & 76.3 & 27.7 & 55.1 & 0.1894 & 0.7098 \\
\midrule
\textbf{\methodname}       & 45.0 & 63.7 & 29.2 & 51.5 & 53.5 & 71.1 & 25.7 & 52.9 & 0.1513 & \underline{0.7507} \\
\bottomrule
\end{tabular} \vspace{-0.1in}
}
\end{table*}
\fi 
\begin{table*}\vspace{-0.1in}
\centering
\caption{Ablation study of loss functions and augmentations evaluated across all training and validation views.}
\label{tab:loss_results}
\resizebox{\textwidth}{!}{
\begin{tabular}{l cc cc cc cc >{\columncolor{blue!6}}c >{\columncolor{blue!6}}c}
\toprule

\multirow{2}{*}{\textbf{Method}} 
& \multicolumn{2}{c}{\textbf{VideoLLaMA3 (\%)}} 
& \multicolumn{2}{c}{\textbf{LLaVA-NeXT (\%)}} 
& \multicolumn{2}{c}{\textbf{LLaVA-OV (\%)}} 
& \multicolumn{2}{c}{\textbf{BLIP-2 (\%)}} 
& \multirow{2}{*}{\textbf{LPIPS $\downarrow$}} 
& \multirow{2}{*}{\textbf{ID Sim $\uparrow$}} \\

\cmidrule(lr){2-3}
\cmidrule(lr){4-5}
\cmidrule(lr){6-7}
\cmidrule(lr){8-9}

& \textbf{IR} & \textbf{MR}
& \textbf{IR} & \textbf{MR}
& \textbf{IR} & \textbf{MR}
& \textbf{IR} & \textbf{MR}
& & \\

\midrule

No Augmentation
& 17.6 & 34.7
& 16.1 & 33.9
& 28.8 & 45.3
& 15.6 & 42.8
& \textbf{0.1438} & \textbf{0.8397} \\

%\addlinespace[3pt]

No Chroma Constraint
& \underline{53.4} & \textbf{72.5}
& \textbf{43.1} & \textbf{66.7}
& \textbf{61.7} & \textbf{80.4}
& \textbf{34.5} & \textbf{61.9}
& 0.1911 & 0.5695 \\

No Landmark Consis.
& \textbf{54.6} & \underline{72.2}
& \textbf{43.1} & \underline{66.0}
& \underline{59.7} & \underline{77.5}
& \underline{28.8} & \underline{56.1}
& \underline{0.1512} & 0.6184 \\

%\addlinespace[3pt]

No Clamp
& 44.6 & 65.0
& 31.5 & 53.5
& 57.0 & 76.3
& 27.7 & 55.1
& 0.1894 & 0.7098 \\

\midrule

\rowcolor{green!6}
\textbf{\methodname}
& 45.0 & 63.7
& 29.2 & 51.5
& 53.5 & 71.1
& 25.7 & 52.9
& 0.1513 & \underline{0.7507} \\

\bottomrule
\end{tabular} \vspace{-0.5in}
}
\end{table*}
\vspace{-0.2in}

\subsection{Ablation Studies}

\textbf{Impact of Losses and Augmentation.} In Table \ref{tab:loss_results} we conduct an ablation study analyzing the impact of our loss functions and augmentations. Removing the data augmentations leads to a drastic drop in both Injection Rate and Mismatch Rate across all VLMs. For instance, causing a 60.9\% relative drop in VideoLLaMA3 Injection Rate and a 46.2\% relative drop in LLaVA-OV Injection Rate. This demonstrates that augmentations are strictly necessary to prevent the adversarial shell from overfitting to specific rendered views. Furthermore, removing the chroma loss ($\mathcal{L}_{\mathrm{chroma}}$) or the landmark loss ($\mathcal{L}_{\mathrm{landmarks}}$) significantly degrades the incredibly important perceptual quality and identity similarity of the rendered images compared to the source images. Quantitatively, disabling the chroma loss causes a 24.1\% relative drop in identity similarity and a 26.3\% decrease in perceptual similarity. Similarly, disabling the landmark loss drops identity similarity by 17.6\%. This confirms that the face-aware regularization mechanisms are vital for achieving the optimal balance between a high rate of semantic transfer and identity preservation. Finally, omitting dynamic opacity clamping (``No Clamp") yields marginal semantic gains, such as a relative 7.9\% increase in LLaVA-NeXT Injection Rate, but introduces unacceptable visual artifacts, causing a 25.2\% relative reduction in perceptual similarity and a 5.4\% drop in identity similarity. %These findings further the idea that clamping the opacity over time is necessary to maintain strong perceptual similarity and to preserve the identity of the source face.
\vspace{-0.1pt}
\iffalse
\begin{table*}
\centering
\caption{Quantitative comparison of our FaceShell3D model against variants with only a single visual encoder utilized during training.}
\label{tab:encoder_results}
\resizebox{\textwidth}{!}{
\begin{tabular}{l cc cc cc cc cc}
\toprule
\multirow{2}{*}{Method} & \multicolumn{2}{c}{VideoLLaMA3 (\%)} & \multicolumn{2}{c}{LLaVA-NeXT (\%)} & \multicolumn{2}{c}{LLaVA-OV (\%)} & \multicolumn{2}{c}{BLIP-2 (\%)} & \multirow{2}{*}{LPIPS $\downarrow$} & \multirow{2}{*}{ID Sim $\uparrow$} \\
\cmidrule(lr){2-3} \cmidrule(lr){4-5} \cmidrule(lr){6-7} \cmidrule(lr){8-9}
& Inj & Mis & Inj & Mis & Inj & Mis & Inj & Mis & & \\
\midrule
\methodname~(SigLIP)          & \underline{42.6} & \underline{61.4} & 19.0 & 39.2 & \underline{53.0} & \underline{70.9} & 20.4 & 47.9 & \underline{0.1484} & \textbf{0.7830} \\
\methodname~(Clip)            & 21.6 & 39.7 & \underline{25.5} & \underline{45.7} & 18.9 & 35.4 & \underline{21.1} & \underline{48.8} & \textbf{0.1306} & \underline{0.7789} \\
\midrule
\textbf{\methodname}       & \textbf{45.0} & \textbf{63.7} & \textbf{29.2} & \textbf{51.5} & \textbf{53.5} & \textbf{71.1} & \textbf{25.7} & \textbf{52.9} & 0.1513 & 0.7507 \\
\bottomrule
\end{tabular}\vspace{-0.1in}
}
\end{table*}
\fi
\begin{table*}\vspace{-0.1in}
\centering
\caption{Quantitative comparison of \methodname\ against variants trained using a single visual encoder.}\vspace{-0.05in}
\label{tab:encoder_results}
\resizebox{\textwidth}{!}{
\begin{tabular}{l cc cc cc cc >{\columncolor{blue!6}}c >{\columncolor{blue!6}}c}
\toprule

\multirow{2}{*}{\textbf{Method}} 
& \multicolumn{2}{c}{\textbf{VideoLLaMA3 (\%)}} 
& \multicolumn{2}{c}{\textbf{LLaVA-NeXT (\%)}} 
& \multicolumn{2}{c}{\textbf{LLaVA-OV (\%)}} 
& \multicolumn{2}{c}{\textbf{BLIP-2 (\%)}} 
& \multirow{2}{*}{\textbf{LPIPS $\downarrow$}} 
& \multirow{2}{*}{\textbf{ID Sim $\uparrow$}} \\

\cmidrule(lr){2-3}
\cmidrule(lr){4-5}
\cmidrule(lr){6-7}
\cmidrule(lr){8-9}

& \textbf{IR} & \textbf{MR}
& \textbf{IR} & \textbf{MR}
& \textbf{IR} & \textbf{MR}
& \textbf{IR} & \textbf{MR}
& & \\

\midrule

\methodname~(SigLIP)
& \underline{42.6} & \underline{61.4}
& 19.0 & 39.2
& \underline{53.0} & \underline{70.9}
& 20.4 & 47.9
& \underline{0.1484} & \textbf{0.7830} \\

\methodname~(CLIP)
& 21.6 & 39.7
& \underline{25.5} & \underline{45.7}
& 18.9 & 35.4
& \underline{21.1} & \underline{48.8}
& \textbf{0.1306} & \underline{0.7789} \\

\midrule

\rowcolor{green!6}
\textbf{\methodname}
& \textbf{45.0} & \textbf{63.7}
& \textbf{29.2} & \textbf{51.5}
& \textbf{53.5} & \textbf{71.1}
& \textbf{25.7} & \textbf{52.9}
& 0.1513 & 0.7507 \\

\bottomrule
\end{tabular}\vspace{-0.1in}
}
\end{table*}
\vspace{-0.00in}

\noindent \textbf{Multi-Encoder Supervision.} Additionally, Table \ref{tab:encoder_results} highlights the benefit of utilizing multiple visual encoders during the feature alignment step. Utilizing multiple encoders in our training architecture produces more effective overall semantic attribute shifting across the black-box VLMs than using either SigLIP or CLIP in isolation. For example, ensembling the models yields a \textcolor{ForestGreen}{+108.3\%} relative increase in VideoLLaMA3 Injection Rate compared to using CLIP alone, and a \textcolor{ForestGreen}{+26.0\%} relative increase in BLIP-2 Injection Rate compared to using SigLIP alone. This demonstrates that ensembling diverse representation spaces during optimization prevents the Gaussian shell from overfitting to a single model's feature space, ultimately yielding more robust and transferrable adversarial perturbations.

\vspace{-0.1in}

\section{Conclusion}
\vspace{-0.1in}
In this work, we introduced \methodname, a novel privacy defense framework designed to protect 3D face avatars from unauthorized semantic inference by VLMs. Our method remains robust through multiple rendered viewpoints by operating directly in the 3D Gaussian space and successfully transfers face attribute semantics without significantly degrading visual clarity or destroying the identity of the source face. By introducing a heavily regularized, learnable auxiliary Gaussian shell, we successfully inject perturbations that manipulate VLM understanding while leaving the underlying baseline geometry and identity-defining appearance completely untouched. 

Our extensive evaluations across multiple state-of-the-art black-box VLMs demonstrate that \methodname~achieves highly competitive targeted attribute transfer and mismatch rates. Crucially, it accomplishes this while maintaining significantly higher perceptual quality and identity similarity than current state-of-the-art 2D semantic manipulation baselines. Ablation studies further validate that our multi-attribute synthesis, multi-view adversarial training, and face-aware regularization mechanisms are strictly essential for achieving robust, view-consistent semantic steering.

While our current framework successfully defends static 3D face avatars, a promising direction for future work involves extending this mechanism into the 4D Gaussian Splatting space. Adapting our learnable auxiliary shell to maintain temporal consistency could safeguard fully animatable and dynamic portrait avatars \cite{cap4d} against continuous semantic inference in video formats.

\vspace{-0.11in}

\section*{Acknowledgments}
\vspace{-0.05in}
This material is based upon work supported by the Center for Identification Technology Research and the National Science Foundation under Grant No. 2601332. This work is also supported in part by the National Science Foundation (IIS-2245652).  We would also like to thank Charlotte Vision Lab members for our valuable discussions.
\bibliographystyle{splncs04}
\bibliography{main}

\newpage

\appendix

\section{Appendix}

In Section \ref{ablation}, we provide additional ablation studies, detailing the performance impact of calculating our landmark consistency loss directly in the 3D space and analyzing the effect of varying the total number of initialized auxiliary shell Gaussians. Furthermore, we showcase the robustness of \methodname to postprocessing, results for \methodname on an additional dataset with a new 3D reconstruction pipeline, failure case analysis, and a visualization of our average noise intensity. Finally, in Section \ref{user} we present a user study which provides further insight into the quality of our rendered images compared to other selected methods.

\subsection{Additional Ablation Studies}
\label{ablation}

\begin{table*}[h]
\centering
\caption{Ablation study comparing \methodname\ with a variant where the landmark loss is computed in 3D space. Results are reported across frontal face views.}
\label{tab:ablation_3dlm}
\resizebox{\textwidth}{!}{
\begin{tabular}{l cc cc cc cc >{\columncolor{blue!6}}c >{\columncolor{blue!6}}c}
\toprule
\multirow{2}{*}{\textbf{Method}} 
& \multicolumn{2}{c}{\textbf{VideoLLaMA3 (\%)}} 
& \multicolumn{2}{c}{\textbf{LLaVA-NeXT (\%)}} 
& \multicolumn{2}{c}{\textbf{LLaVA-OV (\%)}} 
& \multicolumn{2}{c}{\textbf{BLIP-2 (\%)}} 
& \multirow{2}{*}{\textbf{LPIPS $\downarrow$}} 
& \multirow{2}{*}{\textbf{ID Sim $\uparrow$}} \\

\cmidrule(lr){2-3} \cmidrule(lr){4-5} \cmidrule(lr){6-7} \cmidrule(lr){8-9}

& \textbf{IR} & \textbf{MR} 
& \textbf{IR} & \textbf{MR} 
& \textbf{IR} & \textbf{MR} 
& \textbf{IR} & \textbf{MR} 
& & \\

\midrule

\rowcolor{green!6}
\textbf{\methodname}                 
& \textbf{48.7} & 66.4  
& 31.3 & 53.0  
& 56.9 & 74.4  
& 26.7 & 53.7  
& \textbf{0.1499} & \textbf{0.7629} \\

\methodname\ $-$ 3D LM         
& 48.6 & \textbf{67.3}  
& \textbf{36.1} & \textbf{56.9}  
& \textbf{58.1} & \textbf{76.4}  
& \textbf{27.3} & \textbf{54.4}  
& 0.1532 & 0.7342 \\

\bottomrule
\end{tabular}
}
\end{table*}

Table \ref{tab:ablation_3dlm} explores an alternative formulation of our face-aware regularization by shifting the landmark loss calculation from the 2D rendered space directly into the 3D Gaussian parameter space. To achieve this, we first create 2D facial landmark masks (covering the eyes, nose, and lips) for the source face renders. We then map these 2D masks to the 3D auxiliary shells via a perspective projection pipeline:

\begin{enumerate}
    \item \textbf{Perspective Projection:} The 3D spatial means of the shell Gaussians are converted to homogeneous coordinates and multiplied by the camera's full projection matrix. We apply perspective division to transform these points into Normalized Device Coordinates (NDC) bounded between -1.0 and 1.0.
    \item \textbf{Image Space Mapping:} The NDC coordinates are scaled by the camera's resolution to calculate exact 2D continuous pixel coordinates $(u, v)$ in the image space.
    \item \textbf{Bounds Checking:} We strictly filter out any Gaussians that project behind the camera plane or fall outside the 2D image boundaries.
    \item \textbf{Mask Intersection:} For the remaining valid projections, we sample the 2D MediaPipe mask at the discrete $(u, v)$ pixel indices. If a shell Gaussian projects onto an active landmark pixel in any of the training views, it is permanently flagged as a ``\textit{landmark shell}''.
\end{enumerate}

Instead of penalizing 2D pixel distortions, this variant computes the loss directly on the flagged 3D shells by severely penalizing their opacity and explicit color deviations from the baseline. While this direct 3D-space landmark loss slightly improves the accuracy metrics across all evaluated models, for instance, increasing the LLaVA-NeXT\cite{llavanext} injection rate by 7.4\%, and the BLIP-2 injection rate by 2.2\%, it ultimately degrades perceptual fidelity as well. Identity similarity is degraded by 3.8\% and LPIPS is degraded by 2.2\%. Ultimately because of the importance of identity preservation and image fidelity in this task it remains best to perform the landmark loss in the 2D space.

Table \ref{tab:ablation_scale} analyzes the structural and semantic impact of initializing different quantities of auxiliary shell Gaussians. We compare our standard configuration of 50,000 Gaussians against variants with 25,000 and 75,000 Gaussians. The results indicate that increasing the number of shell Gaussians yields a significant accuracy improvement that definitively warrants the small trade-off in visual fidelity. For example, scaling the auxiliary shells from 50,000 to 75,000 provides rather large improvements in semantic manipulation. We observe a 7.0\% relative increase in the injection rate on VideoLLaMA3\cite{VL3} and a 14.1\% relative increase in the LLaVA-NeXT\cite{llavanext} mismatch rate among improvements in all other accuracy related metrics as well. These substantial semantic gains easily justify the very minor penalty incurred with LPIPS and identity similarity degrading by only 1.9\% and 3.2\% respectively.

\begin{table}[H]
\centering
\caption{Ablation study evaluating the impact of varying the number of shell Gaussians compared to our standard configuration. Results are reported across frontal face views.}
\label{tab:ablation_scale}
\resizebox{\textwidth}{!}{
\begin{tabular}{l cc cc cc cc >{\columncolor{blue!6}}c >{\columncolor{blue!6}}c}
\toprule

\multirow{2}{*}{\textbf{Method}} 
& \multicolumn{2}{c}{\textbf{VideoLLaMA3 (\%)}} 
& \multicolumn{2}{c}{\textbf{LLaVA-NeXT (\%)}} 
& \multicolumn{2}{c}{\textbf{LLaVA-OV (\%)}} 
& \multicolumn{2}{c}{\textbf{BLIP-2 (\%)}} 
& \multirow{2}{*}{\textbf{LPIPS $\downarrow$}} 
& \multirow{2}{*}{\textbf{ID Sim $\uparrow$}} \\

\cmidrule(lr){2-3} 
\cmidrule(lr){4-5} 
\cmidrule(lr){6-7} 
\cmidrule(lr){8-9}

& \textbf{IR} & \textbf{MR} 
& \textbf{IR} & \textbf{MR} 
& \textbf{IR} & \textbf{MR} 
& \textbf{IR} & \textbf{MR} 
& & \\

\midrule

Ours (25k Gaussians)        
& 40.7 & 59.0  
& 23.1 & 41.6  
& 50.0 & 67.3  
& 25.0 & 51.3  
& \textbf{0.1466} & \textbf{0.8003} \\

\rowcolor{green!6}
\textbf{Ours (50k Gaussians)}                 
& 48.7 & 66.4 
& 31.3 & 53.0 
& 56.9 & 74.4 
& \underline{26.7} & 53.7  
& \underline{0.1499} & \underline{0.7629} \\

Ours (75k Gaussians)         
& \underline{52.1} & \underline{70.1}  
& \underline{35.7} & \underline{58.3}  
& \underline{61.3} & \underline{79.4}  
& 26.6 & \textbf{54.7}  
& 0.1527 & 0.7388 \\

Ours (100k Gaussians)         
& \textbf{52.3} & \textbf{71.6}  
& \textbf{38.4} & \textbf{60.6}  
& \textbf{61.7} & \textbf{80.0}  
& \textbf{26.9} & \underline{54.1}  
& 0.1527 & 0.7295 \\

\bottomrule
\end{tabular}
}
\end{table}

\vspace{-0.2in}
\begin{table}[H]
\centering
\caption{Comparison of robustness between \methodname and VAttack\cite{vattack}}
\label{tab:robustness}
\resizebox{0.75\columnwidth}{!}{
\begin{tabular}{lc cccc}
\toprule
\textbf{Condition} & \textbf{Method} & \textbf{Inject} $\uparrow$ & \textbf{MisMatch} $\uparrow$ & \textbf{LPIPS} $\downarrow$ & \textbf{ID Sim} $\uparrow$ \\
\midrule
\multirow{3}{*}{Pre-Transformation}
  & GT                    & 5.0\%  & 17.9\% & 0.0000 & 1.0000 \\
  & VAttack\cite{vattack}& 44.4\% & \textbf{74.0\%} & 0.4039 & 0.5255 \\
  & \textbf{Ours}                  & \textbf{48.7\%} & 66.4\% & \textbf{0.1499} & \textbf{0.7629} \\
\cmidrule(lr){1-6}
\multirow{3}{*}{JPEG ($Q=15$)}
  & GT                    & 5.1\%  & 19.4\% & 0.0756 & 0.9268 \\
  & VAttack\cite{vattack}& 21.1\% & 49.3\% & 0.2767 & 0.5875 \\
  & \textbf{Ours}                  & \textbf{37.4\%} & \textbf{58.1\%} & \textbf{0.2249} & \textbf{0.6174} \\
\cmidrule(lr){1-6}
\multirow{3}{*}{Resize ($196\times196$)}
  & GT                    & 5.4\%  & 20.1\% & 0.0419 & 0.9806 \\
  & VAttack\cite{vattack}& 35.7\% & \textbf{65.9\%} & 0.3769 & 0.5506 \\
  & \textbf{Ours}                  & \textbf{37.0\%} & 58.3\% & \textbf{0.2074} & \textbf{0.6213} \\
\cmidrule(lr){1-6}
\multirow{3}{*}{Screen Artifact}
  & GT                    & 5.4\%  & 18.3\% & 0.1750 & 0.8655 \\
  & VAttack\cite{vattack}& 24.7\% & 51.9\% & \textbf{0.3606} & \textbf{0.6367} \\
  & \textbf{Ours}                  & \textbf{47.1\%} & \textbf{65.9\%} & 0.3748 & 0.6339 \\
\cmidrule(lr){1-6}
\multirow{2}{*}{Re-rendered}
  & GT                    & 5.0\%  & 17.4\% & 0.0467 & 0.9581 \\
  & \textbf{Ours}                  & \textbf{35.1\%} & \textbf{54.3\%} & \textbf{0.1713} & \textbf{0.7801} \\
\bottomrule
\end{tabular}
}
\end{table}

Table \ref{tab:robustness} evaluates \methodname~under 4 real-world distortions:
(i)~JPEG compression (15\% quality);
(ii)~resizing to $196\!\times\!196$;
(iii)~a screen-photography simulation pipeline comprising Gaussian blur, contrast/brightness scaling, a high-frequency grid overlay, and additive Gaussian noise; and
(iv)~rendering-level perturbations via exposure modulation, color temperature shifts, and Gaussian scale variation.
Across all settings, \methodname~shows consistent robustness, outperforming the 2D SoTA VAttack\cite{vattack}.

\begin{table}[H]\vspace{-0.15in}
\centering
\caption{Attack results on APPA-REAL\cite{appareal} with 3D avatars constructed via LGM\cite{lgm}.}
\label{tab:cross_model_results}
\resizebox{\columnwidth}{!}{
\begin{tabular}{ll cccc @{\hspace{1em}} cc}
\toprule
& & \multicolumn{4}{c}{\textbf{Target VLM}} & \multicolumn{2}{c}{\textbf{Perceptual}} \\
\cmidrule(lr){3-6} \cmidrule(l){7-8}
\textbf{Method} & & \textbf{VLLaMA3} & \textbf{LL-Next} & \textbf{LL-OV} & \textbf{BLIP-2} & \textbf{LPIPS} $\downarrow$ & \textbf{ID Sim} $\uparrow$ \\
\midrule
\multirow{2}{*}{Original} 
& IR $\uparrow$ & 3.6\% & 4.3\% & 5.0\% & 12.9\% & \multirow{2}{*}{0.000} & \multirow{2}{*}{1.000} \\
& MR $\uparrow$ & 17.9\% & 21.4\% & 25.0\% & 38.6\% & & \\
\midrule
\multirow{2}{*}{VAttack\cite{vattack}} 
& IR $\uparrow$ & 55.0\% & 49.3\% & 55.7\% & 39.3\% & \multirow{2}{*}{0.4121} & \multirow{2}{*}{0.4257} \\
& MR $\uparrow$ & 84.3\% & 70.7\% & 83.6\% & 72.9\% & & \\
\midrule
\multirow{2}{*}{\textbf{\methodname}} 
& \textbf{IR} $\uparrow$ & \textbf{50.7\%} & \textbf{29.0\%} & \textbf{46.9\%} & \textbf{21.8\%} & \multirow{2}{*}{\textbf{0.153}} & \multirow{2}{*}{\textbf{0.801}} \\
& \textbf{MR} $\uparrow$ & \textbf{73.6\%} & \textbf{52.9\%} & \textbf{72.5\%} & \textbf{47.0\%} & & \\
\bottomrule
\end{tabular}
}
\end{table}

Table \ref{tab:cross_model_results} reports results on 20 identities from APPA-REAL \cite{appareal}, a dataset of user-uploaded (non-celebrity) faces, reconstructed via LGM\cite{lgm}, a general-purpose image-to-3D model.
\methodname~maintains strong performance across this real-world dataset and alternative 3D reconstruction pipeline, demonstrating generalization beyond the evaluation setting of the main paper.

\begin{table}[H]
    \centering
    \footnotesize 
    \renewcommand{\arraystretch}{0.9} 
    \caption{Count of samples exhibiting each possible number of mismatched and injected attributes. All results come from evaluation on VideoLLaMA3\cite{VL3}}
    \label{tab:rate_distributions_full}
    % The star (*) and \columnwidth force it to span the full width.
    % @{\extracolsep{\fill}} evenly spaces the columns.
    \scalebox{0.7}{
    \begin{tabular*}{\columnwidth}{@{\extracolsep{\fill}} l *{8}{c} @{}}
        \toprule
        \textbf{Metric} & \textbf{0} & \textbf{1} & \textbf{2} & \textbf{3} & \textbf{4} & \textbf{5} & \textbf{6} & \textbf{7} \\
        \midrule
        Mismatch & 0 & 2 & 4 & 17 & 16 & 33 & 21 & 7 \\
        Injection  & 1 & 6 & 20 & 25 & 27 & 15 & 5 & 1 \\
        \bottomrule
    \end{tabular*}}
\end{table}

To further analyze failure cases Table \ref{tab:rate_distributions_full} details the number samples achieving each possible injection and mismatch count (0-7 attributes). \methodname~exhibits near-zero failure: mismatch occurs in all 100 samples, and target attribute injection fails completely in only 1 case demonstrating consistent adversarial effectiveness across the attribute space.

\begin{figure}[H]
    \centering
    % \linewidth scales the image to perfectly fit the single column width
    \includegraphics[width=0.95\linewidth]{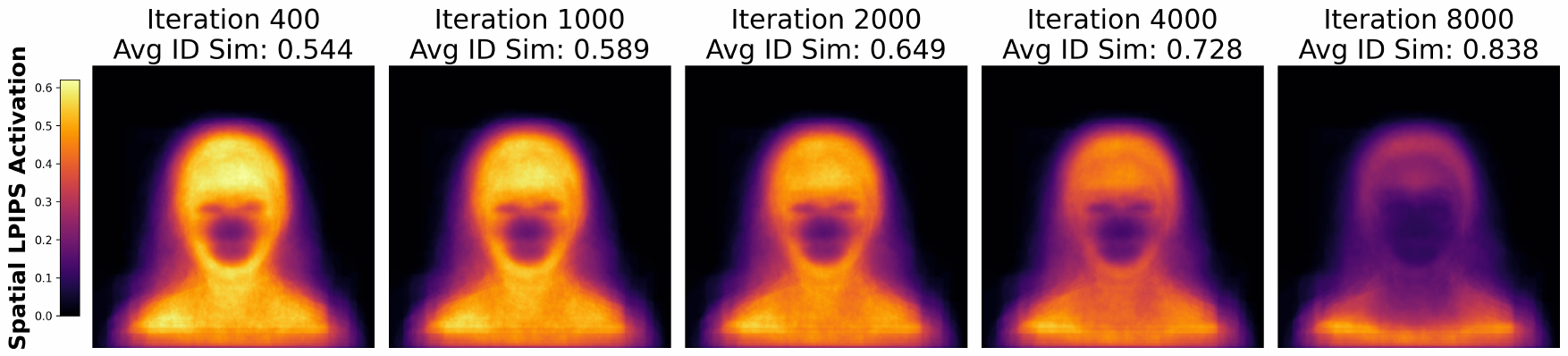}
    
    \caption{Progression of spatial LPIPS activation averaged across 100 subjects.}
    \label{fig:heatmap_progression}
\end{figure}

Figure \ref{fig:heatmap_progression} plots spatial LPIPS and identity similarity averaged across all samples at varying optimization iterations. Early iterations exhibit broad structural distortion; late iterations converge to globally minimized perturbation. At intermediate stages, \methodname~concentrates perturbations in semantically non-critical facial regions, disrupting VLM interpretation while enforcing low LPIPS deviation over identity-critical areas, thus preserving subject identity throughout optimization.

\subsection{User Study}
\label{user}

\begin{table}[h]
\centering
\caption{Results from our user study. Ranks are relative ranging from 1st (best) to 4th (worst).}
\label{tab:user_study}
\begin{tabular}{l r r}
\toprule
\textbf{Method} & \textbf{Avg. Identity Preservation $\downarrow$} & \textbf{Avg. Perceptual Quality $\downarrow$} \\
\midrule
AttackVLM\cite{attackvlm}      & \underline{2.42}  & 2.74  \\
M-Attack\cite{mattack}       & 2.85  & \underline{2.67}  \\
V-Attack\cite{vattack}       & 3.13  & 3.26  \\
\midrule
\textbf{\methodname}        & \textbf{1.61}  & \textbf{1.33}  \\
\bottomrule
\end{tabular}
\end{table}

Table \ref{tab:user_study} presents a user study which compares our method with three others. M-Attack\cite{mattack} and V-Attack\cite{vattack} were chosen for their strong injection rate and mismatch rate across a variety of models while AttackVLM\cite{attackvlm} was chosen for its strong performance in terms of identity preservation and LPIPS. This study was conducted with 13 participants who were each presented with 10 sets of images. Each set contained a source face image alongside a noised version of that source face generated by each of the four methods. The names of each method were anonymized and the position of each method was randomized for each set. Participants were asked to rank each method relatively from 1st (best) to 4th (worst) in two metrics for each set of images. The first metric was the identity preservation of the noised image and the second was the perceptual quality of the noised image. Our results show that users strongly agree that the results generated by \methodname are consistently much better in terms of perceptual quality and identity preservation compared to the other methods provided. Our method achieved a dominant average rank of 1.33 for perceptual quality, outperforming the next best method, M-Attack\cite{mattack}, at 2.67, and an average rank of 1.61 for identity preservation, significantly leading the next best method, AttackVLM\cite{attackvlm}, at 2.42.

\end{document}